\documentclass[11pt,letterpaper,teaser]{planstyle}
\usepackage[T1]{fontenc}
\usepackage{microtype}           
\usepackage{nicefrac}            
\usepackage{xspace}              
\usepackage{amsmath, amssymb, amsfonts, amsthm, mathtools, mathrsfs}
\usepackage{dsfont}              
\usepackage{fdsymbol}            

\usepackage{booktabs}            
\usepackage{nicematrix}         
\usepackage{multirow}
\usepackage{makecell}
\usepackage{bigstrut}
\usepackage{enumitem}            
\setitemize{label=\textbullet, leftmargin=*, nolistsep}
\usepackage{graphicx}
\usepackage{adjustbox}           
\usepackage{float}               
\usepackage{wrapfig}             
\usepackage{subcaption}          
\expandafter\def\csname ver@subfig.sty\endcsname{}  
\usepackage[dvipsnames]{xcolor}    
\usepackage{color}
\usepackage{fontawesome5}        
\usepackage{pifont}              
\usepackage{hyperref}            
\hypersetup{colorlinks=true, citecolor=LightBlue}
\usepackage[all]{hypcap}         
\usepackage{cleveref}            
\usepackage{url}                 
\usepackage{algorithm}
\usepackage{algorithmicx}
\usepackage{algpseudocode}
\usepackage[normalem]{ulem}      
\usepackage{lipsum}              
\usepackage{rotating}            
\usepackage{stackengine}         
\usepackage{caption}             
\usepackage{listings}            
\usepackage{soul}                
\usepackage{gradient-text}       
\usepackage{pgfplots}            
\pgfplotsset{compat=newest}
\usepackage{pgfplotstable}       
\usepackage{tikz}                
\usepackage{svg}                 

\usepackage[comma,numbers,sort,compress]{natbib}
\definecolor{demphcolor}{RGB}{125,125,125}    

\hypersetup{
    colorlinks = true,
    citecolor = {YaleBlue},
}
\tcbuselibrary{breakable, skins}
\newtcolorbox{planbox}[1]{
  enhanced,
  breakable,
  colback=white,
  colframe=IllinoisOrange!80,
  coltitle=IllinoisBlue,
  fonttitle=\bfseries\sffamily,
  title=#1,
  titlerule=0.8pt,
  boxrule=1pt,
  left=3mm, right=3mm, top=2mm, bottom=2mm,
  boxsep=1mm,
  before upper=\smallskip,
}

\newcommand{\eg}{\textit{e.g.},\xspace}      

\crefname{equation}{Eq.}{Eqs.}
\crefformat{section}{\S#2#1#3}
\crefformat{subsection}{\S#2#1#3}
\crefformat{subsubsection}{\S#2#1#3}
\crefrangeformat{section}{\S\S#3#1#4 to~#5#2#6}
\crefmultiformat{section}{\S\S#2#1#3}{ and~#2#1#3}{, #2#1#3}{ and~#2#1#3}

\usepackage{tcolorbox}
\tcbuselibrary{listings,breakable}
\usepackage{cuted} 
\usepackage{dblfloatfix}
\usepackage{hyperref}
\usepackage{url}
\usepackage{wrapfig}
\usepackage{tcolorbox}
\usepackage{pgfplots}
\usepackage{wrapfig}
\pgfplotsset{compat=1.18}
\definecolor{saprE}{RGB}{201, 182, 228} 
\definecolor{rlE}{RGB}{245,130,55}
\definecolor{fire0}{HTML}{FFF2B2}
\definecolor{fire1}{HTML}{C1121F}
\definecolor{fire2}{HTML}{FFB347}
\definecolor{fire3}{HTML}{FF8A3D}
\definecolor{fire4}{HTML}{FF4D00}
\definecolor{fire5}{HTML}{FF4A3A}
\definecolor{fire6}{HTML}{CE0A18}

\definecolor{CUBlue}{HTML}{4285F4}
\definecolor{Google}{HTML}{EA4335}

\definecolor{pyraAmber}{HTML}{FFB347}
\definecolor{pyraOrange}{HTML}{FF6B3D}
\definecolor{pyraRed}{HTML}{E63946}

\definecolor{color1}{RGB}{220, 237, 220}  
\definecolor{color2}{RGB}{252, 224, 225}  
\definecolor{color3}{RGB}{220, 235, 247}  
\definecolor{color4}{RGB}{230, 230, 230}  

\theoremstyle{plain}

\theoremstyle{definition}

\theoremstyle{remark}

\usepackage{enumitem}
\setlist[itemize]{leftmargin=*}

\usepackage{gradient-text}
\definecolor{IllinoisBlue}{HTML}{13294B}
\definecolor{IllinoisOrange}{HTML}{FF5F05}
\definecolor{ForestGreen}{RGB}{34,139,34}
\definecolor{CustomPurple}{HTML}{884AB2}

\newcommand{\modelnamenc}{{CogniRoute}}
\newcommand{\benchmarknamenc}{{\textsc{OmniSocialBench}}}

\newcommand{\modelnamegradient}{%
\textbf{{\gradientRGB{CogniRoute}{245,130,55}{120,95,240}}\xspace}%
}

\newcommand{\benchmarknamegradient}{%
\textbf{\textsc{\gradientRGB{OmniSocialBench}{80,170,210}{236,112,180}}\xspace}%
}

\title{\vspace{-0.5cm}
\modelnamegradient: Learning to Route Social Evidence in Omni-Modal Models}

\author{
\vspace{-0.3cm}
    \textbf{Yifan Shen\textsuperscript{\textcolor{IllinoisOrange}{1}}, Pei Tian\textsuperscript{\textcolor{CUBlue}{2}}, Xinzhuo Li\textsuperscript{\textcolor{IllinoisOrange}{1}}, Bowen Fang\textsuperscript{\textcolor{CUBlue}{2}}, Shujun Xia\textsuperscript{\textcolor{CUBlue}{2}}, Bingxuan Li\textsuperscript{\textcolor{IllinoisOrange}{1}}, Ana Jojic\textsuperscript{\textcolor{IllinoisOrange}{1}}} \\
    \textbf{Wenming Ye\textsuperscript{\textcolor{Google}{3}}, Xu Cao\textsuperscript{\textcolor{IllinoisOrange}{1}}, James Matthew Rehg\textsuperscript{\textcolor{IllinoisOrange}{1}}, Ismini Lourentzou\textsuperscript{\textcolor{IllinoisOrange}{1}}}  
    }

\affil{\textcolor{IllinoisOrange}{\textsuperscript{1} University of Illinois Urbana-Champaign}, \textcolor{CUBlue}{\textsuperscript{2}Columbia University}, \textcolor{Google}{\textsuperscript{3}Google}
\vspace{-0.1cm}\\
\texttt{\{yifan26,lourent2\}@illinois.edu}
\vspace{-0.5cm}
}

\correspondingauthor{
  \textsuperscript{$\star$}Equal Contribution \\
  \textsuperscript{$\dagger$}Equal Corresponding Authors
}

\begin{document}

\setabstractlogo[9mm]{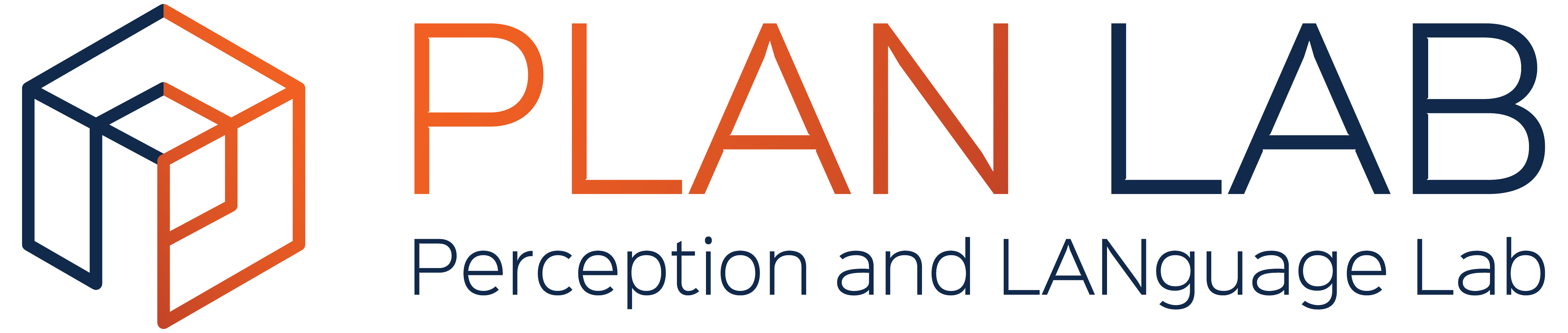}

\begin{abstract}
Omni-modal models can ingest video, audio, and text, but unified access to multiple modalities does
not guarantee that a model uses the right evidence. This gap is especially pronounced in social video
question answering, where the answer may hinge on a gesture, vocal tone, temporal cue, or mismatch
between what is said and what is visually expressed. We introduce \modelnamegradient{}, a schema-guided
Mixture-of-Experts framework for social omni reasoning. \modelnamenc{} uses a training-only
cognitive schema that factorizes each example by cross-modal relation, reasoning demand, and
temporal scope, and aligns global routing signatures with this structure during supervised
fine-tuning. We further introduce route-aware reinforcement learning, which jointly optimizes token
generation and expert allocation using rewards for answer correctness, modality-consistent
reasoning, and cognitive temporal grounding. To support training and evaluation, we construct
\benchmarknamegradient{}, a diagnostic social video QA resource with 118K structured training examples,
grounded reasoning traces, schema labels, temporal evidence spans, and a manually verified
evaluation split. \modelnamenc{} achieves 59.38\% average accuracy on \benchmarknamenc{}, improving
over the strongest proprietary baseline by 15.33 percentage points and the strongest open-source
omni baseline by 26.77 points, with the largest gains on questions requiring audio-visual
coordination, conflict resolution, and temporally grounded social inference.
\vspace{2mm}
\url{https://plan-lab.github.io/cogniroute}
\end{abstract}

\maketitle

\section{Introduction}
\label{sec:intro}
Recent omni models can process video, audio, and text within a single autoregressive system, creating a common interface for multimodal perception and reasoning \cite{xu2025qwen3,xu2025qwen25omnitechnicalreport, comanici2025gemini}.
However, progress in omni modeling has exposed a persistent gap between having access to multiple modalities and using the right evidence for a particular question \cite{park2025assessing,sung2024avhbench}.
This gap is especially pronounced in social video question answering.  Social meaning is often carried by the relation between expression, action, speech content, tone, laughter, silence, and timing. A question may depend on a glance or hand motion, on what is said and how it is said, on a combination of both, or on a conflict between the two.
In such settings, the central challenge extends beyond representing multiple modalities in a shared model to determining which evidence is reliable and where the decisive cue occurs in time.

Existing video QA and audio-visual benchmarks usually specify what the correct answer is \cite{zadeh2019social,liu2024tackling,li2024mvbench,wu2024longvideobench}, but not why a particular modality should be trusted, whether audio and visual evidence are complementary or conflicting, what reasoning operation is required, or where the relevant evidence occurs in time. As a result, a model may produce the correct answer while relying on the wrong modality, attending to an irrelevant temporal segment, or exploiting dataset priors. This limitation is difficult to diagnose from answer accuracy alone and is particularly harmful for social reasoning, where plausible but incorrect interpretations often arise from missing a subtle cue.

Sparse Mixture-of-Experts architectures offer a natural mechanism for conditional computation \cite{riquelme2021scaling,chi2022representation,pan2025fsmoe}, but current MoE routing does not solve this problem by itself. In standard multimodal MoE models, routing is typically driven by local token representations and optimized indirectly through next-token prediction and load balancing. Recent MoE and routing methods improve efficiency \cite{lin2026moe,wu2025routing}, modality specialization \cite{li2025uni,xia2025smar}, or adaptive expert usage \cite{hwang2023tutel,yan2025accelerating}, but they do not provide explicit sample-level supervision that tells the router what kind of multimodal reasoning the input requires. As a result, the router is not explicitly trained to produce routing patterns that distinguish whether a question requires visual evidence, speech content or prosody, audio-visual coordination, cross-modal conflict resolution, temporal linking, or higher-level social inference.\looseness-1

In this work, we use social video question answering as a strong testbed for omnimodal understanding and introduce \modelnamegradient{}, a schema-guided MoE training framework for joint visual and audio reasoning. 
The key idea is to expose the latent evidence structure of each training example through a compact, training-only \textit{Cognitive Schema} that summarizes three sample-level properties: the relation between audio and visual evidence, the reasoning operation required by the question, and the temporal scope of the relevant cue.
The schema is used only during training to shape routing behavior; at inference, \modelnamenc{} receives only the original video, audio, and question, and generates the answer directly.
During supervised fine-tuning, our \textit{Schema-Aligned Predictive Routing} objective encourages routing patterns to become predictive of the sample’s evidence structure, including whether the question depends on visual evidence, audio evidence, audio-visual coordination, cross-modal conflict, temporal reasoning, or higher-level social inference. 

Answer likelihood alone provides limited guidance for optimizing expert allocation, because the same answer can be produced through different internal routes, including routes that ignore the relevant modality or temporal region. \modelnamenc{} therefore introduces \emph{Route-Aware MoE Reinforcement Learning}, which optimizes both token generation and expert routing using rollout-level rewards that combine answer correctness with modality-consistent reasoning and cognitive temporal grounding, so that successful generations are those that not only predict the right answer, but also ground the decision in the appropriate modalities and time span.\looseness-1

Because standard QA datasets do not indicate which visual or audio cues support each answer, how those cues relate, or when they occur, we construct \benchmarknamegradient{}, a social video QA resource with a structured training set and a manually verified benchmark split, enabling controlled evaluation of whether models use the appropriate modality, resolve audio-visual conflict, and ground their answers in the correct temporal region. The training set augments original video-question-answer pairs with evidence summaries, schema labels, grounded reasoning, routing-related annotations, and temporal evidence spans, providing the supervision required to learn evidence-aware routing. The held-out evaluation set is manually verified 
and supports diagnostic analysis across social inference types, including mental-state inference, pragmatic meaning, action goals, and social norms.
To reduce annotation drift, both splits are built with a staged pipeline that first extracts observation-level visual and audio evidence, then assigns schema labels and generates reasoning from this shared evidence while preserving the original answers.

On \benchmarknamenc{}, \modelnamenc{} obtains 59.38\% average accuracy, improving over the strongest proprietary baseline by 15.33 percentage points and over the strongest open-source omni baseline by 26.77 points.
The improvements are especially large on socially grounded categories such as action goal and social-norm inference, where the model must coordinate visual and audio cues, resolve ambiguity or conflict, and ground the interpretation in time.
Beyond \benchmarknamenc{}, \modelnamenc{} improves over the base model on 8 out of 10 public audio-visual and video reasoning benchmarks, demonstrating that the proposed routing and reinforcement objectives transfer beyond the constructed social benchmark. Our contributions are summarized as follows:
\begin{itemize}[itemsep=0.5ex, parsep=0pt, topsep=-2.3pt, leftmargin=0.4cm]
    
    \item We introduce \modelnamegradient{}, a schema-guided MoE training framework for omni social video reasoning that aligns
     expert routing with the evidence structure of each question, including modality relation, reasoning demand, and temporal scope. This enables the model to distinguish when to rely on visual cues, audio cues, audio-visual coordination, cross-modal conflict resolution, or temporally grounded social inference.
    
    \item We propose {Route-Aware MoE Reinforcement Learning} for multimodal MoE reasoning, which jointly optimizes answer generation and expert allocation with rollout-level rewards for answer correctness, modality-consistent reasoning, and cognitive temporal grounding. This encourages the model to produce correct answers through the appropriate modalities and temporal evidence.
        
    \item We introduce \benchmarknamegradient{}, a social video QA resource designed to evaluate evidence-aware omni reasoning. Unlike standard QA benchmarks that primarily supervise answers, \benchmarknamenc{} provides structured training supervision over modality evidence, audio-visual relations, reasoning demands, and temporal evidence spans, together with a manually verified evaluation split for fine-grained diagnostic analysis.\looseness-1
\end{itemize}

\section{Related Work}
\label{sec:relatedwork}

\noindent \textbf{Omnimodal Large Language Models and audio-visual reasoning.} Recent MLLMs have expanded from image-text understanding to omni-modal models that jointly process text, vision, and audio within a unified interaction framework~\cite{bai2025qwen25vltechnicalreport,bai2025qwen3,lin2024video,liu2023visual,chen2024internvl,li2024llava,hurst2024gpt,fu2025vita}. Systems such as Qwen-Omni~\cite{xu2025qwen25omnitechnicalreport,xu2025qwen3} and Gemini~\cite{team2023gemini,team2024gemini,comanici2025gemini} demonstrate strong real-time and holistic audio-visual understanding capabilities. However, unified multimodal access~\cite{xu2025qwen25omnitechnicalreport,xu2025qwen3,comanici2025gemini} does not specify how evidence should be selected or how computation should be allocated for each question. Existing OLM training~\cite{ye2025omnivinci,yu2024salmonn} primarily optimizes generated responses, leaving modality reliability, cross-modal agreement or conflict, temporal grounding, and social evidence use only indirectly supervised. \modelnamenc{} addresses this gap by using sample-level schema supervision to guide MoE routing according to modality relation, reasoning demand, and temporal scope.\looseness-1

\noindent \textbf{MoE-based Multimodal Models.}
Mixture-of-Experts (MoE) models enable conditional computation by routing inputs to a subset of experts~\cite{eigen2013learning}. In multimodal learning, MoE has been used to specialize computation across modalities~\cite{bao2022vlmo,mustafa2022multimodal,chen2024eve,shen2026egoforge}, tasks~\cite{satar2022romeroleawaremixtureofexperttransformer,li2023pace}, and domains through modality- or task-conditioned experts, sparse routing~\cite{shen2024mome}, and parameter-efficient expert adapters~\cite{long2024multiwayadapateradaptinglargescalemultimodal}. These methods improve capacity and efficiency~\cite{li2024aria,lin2026moe}, but their routers are typically optimized through next-token prediction, load balancing, or general efficiency objectives, rather than fine-grained supervision to learn which modality is reliable, how audio and visual cues relate, what reasoning is required, and where the decisive cue occurs.  In contrast, \modelnamenc{} aligns routing signatures with sample-level schema labels and uses route-aware reinforcement learning to optimize expert allocation for correct, modality-consistent, and temporally grounded answers.

\section{Method}
\label{sec:method}
\begin{figure}[!t]
    \centering    
\includegraphics[width=\linewidth]{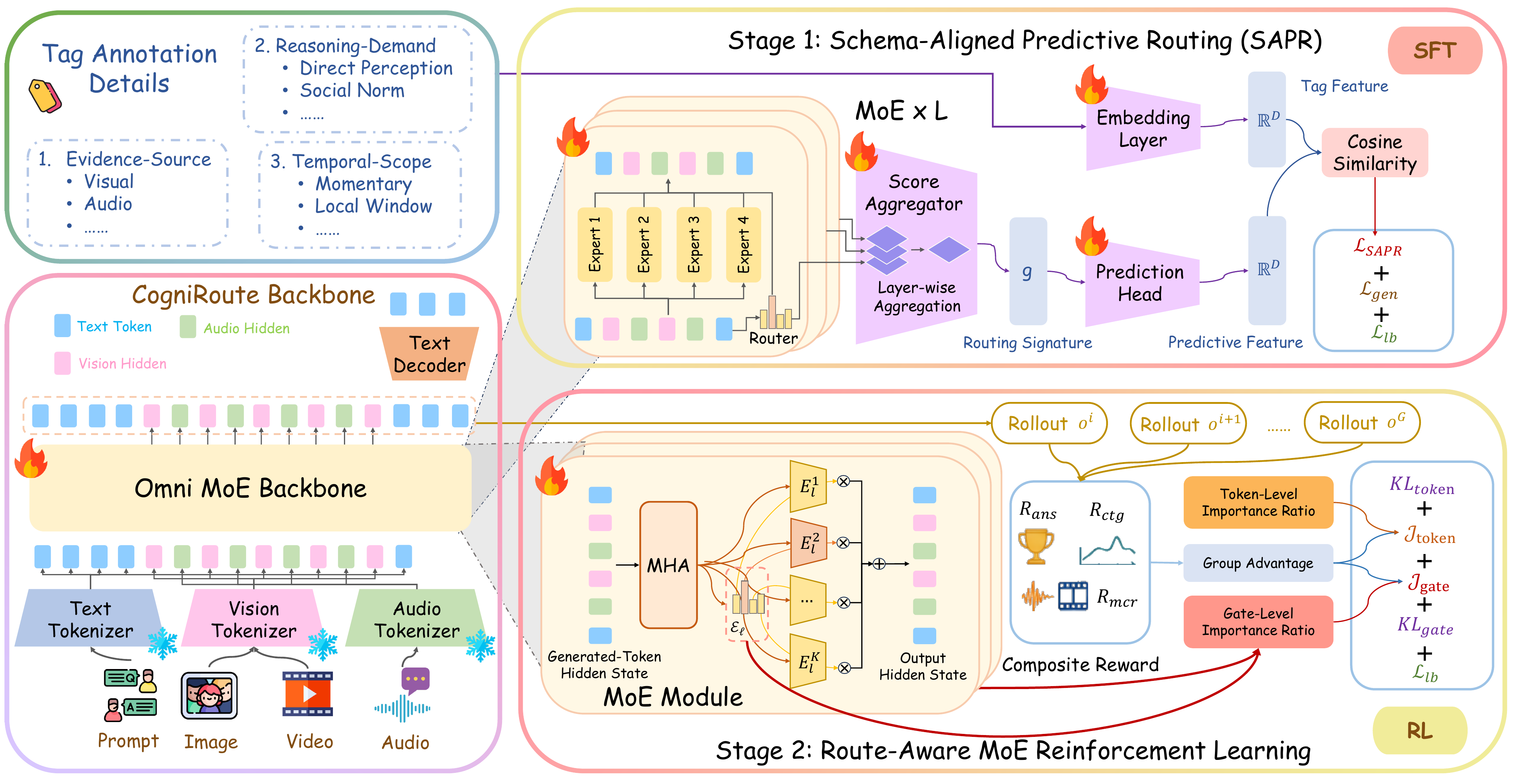}
\caption{\textbf{\modelnamegradient{} optimizes expert routing for omni-modal reasoning.}
SAPR aligns routing with evidence source, reasoning type, and temporal scope, while Route-Aware MoE RL jointly optimizes answer generation and expert allocation.}
    \label{fig:method}
\end{figure}

Given an input $x\!=\!(v,a,q)$ consisting of a video $v$, audio $a$, and question $q$, the model generates an output sequence $o\!=\!(r,y)$, where $r$ is the reasoning trace and $y$ is the final answer. 
Our goal is to train an omni MoE model whose routing behavior reflects the evidence structure of the input: which modality is reliable, how audio and visual cues relate, what reasoning operation is required, and where the decisive cue occurs in time.
\modelnamenc{} has two training stages.
During the supervised fine-tuning stage I, {Schema-Aligned Predictive Routing}  (\Cref{sec:sapr}) aligns global routing signatures with a training-only schema tag set $C$ describing modality relation, reasoning demand, and temporal scope.
In stage II, Route-aware MoE Reinforcement Learning (\Cref{sec:rl}) updates both generated tokens and expert-routing actions using rollout-level rewards for answer correctness, modality-consistent reasoning, and a temporal grounding reward that checks whether the final reasoning state points to the correct time region in the clip (\Cref{sec:ctg}). Figure \ref{fig:method} shows the overview of our method.\looseness-1

\subsection{Schema-Aligned Predictive Routing  (SAPR)}
\label{sec:sapr}
Standard MoE routing is computed from local token representations. This is effective for conditional computation, but it provides little explicit guidance about the sample-level evidence structure needed for social reasoning. Two questions with similar answers may require very different internal computation: one may depend on facial motion, another on vocal tone, and another on the relation or conflict between speech and visual behavior. We therefore introduce \textit{Schema-Aligned Predictive Routing} (SAPR), an auxiliary supervised objective that encourages global routing patterns to encode the evidence structure of each sample.

For input token $i$ at MoE layer $l$, the router outputs a dense pre-top-$k$ probability distribution
\( 
\mathbf{p}_{i,l}=\rho_{\theta}^{l}(\mathbf{h}_{i,l}) \in \Delta^{N-1},
\)
where $\mathbf{h}_{i,l}$ is the hidden state, $N$ is the number of experts, and $\Delta^{N-1}$ denotes the probability simplex. We use the dense distribution before top-$k$ truncation so that the alignment signal can reach all experts, not only the selected ones. Since the goal is to align routing with the task structure already present in the input, the routing signature is built only from input tokens and does not use generated reasoning tokens.

Let $\mathcal{I}_v$, $\mathcal{I}_a$, and $\mathcal{I}_q$ denote the input-token indices for video, audio, and question tokens. For each layer $l$ and modality $m\in\{v,a,q\}$, we compute the modality-pooled routing distribution
\begin{equation}
\setlength{\abovedisplayskip}{3pt}
\setlength{\belowdisplayskip}{3pt}
\setlength{\abovedisplayshortskip}{2pt}
\setlength{\belowdisplayshortskip}{2pt}
\bar{\mathbf{p}}^{m}_{l}
=
\frac{1}{|\mathcal{I}_m|}
\sum_{i\in \mathcal{I}_m}
\mathbf{p}_{i,l},
\qquad
\bar{\mathbf{p}}^{m}_{l}\in\mathbb{R}^{N}.
\end{equation}
Pooling by modality prevents long audio or video streams from overwhelming the question tokens, while preserving modality-specific routing statistics. Because expert indices are layer-specific, we keep routing statistics layerwise and project them into a shared space:
\begin{equation}
\setlength{\abovedisplayskip}{3pt}
\setlength{\belowdisplayskip}{3pt}
\setlength{\abovedisplayshortskip}{2pt}
\setlength{\belowdisplayshortskip}{2pt}
\mathbf{g}
=
\frac{1}{L}
\sum_{l=1}^{L}
P_l\,
\operatorname{Concat}
\left(
\bar{\mathbf{p}}^{v}_{l},
\bar{\mathbf{p}}^{a}_{l},
\bar{\mathbf{p}}^{q}_{l}
\right)
\in \mathbb{R}^{D},
\end{equation}
where $P_l\in\mathbb{R}^{D\times 3N}$ is a learnable projection for layer $l$. The resulting vector $\mathbf{g}$ is the global routing signature of the sample.
Each training sample is accompanied by a training-only schema tag set \(\mathcal{C}\)
where the tags describe the sample's cross-modal evidence relation, reasoning demand, and temporal scope, respectively. We describe the tag taxonomy and annotation pipeline in \cref{sec:data}.
We then convert the schema tags  \(\mathcal{C}\) into a target embedding. Let $\mathcal{V}_{\mathrm{tag}}$ be the global schema vocabulary and let $W_{\mathrm{tag}}\in\mathbb{R}^{|\mathcal{V}_{\mathrm{tag}}|\times D}$ be a learnable tag embedding matrix. For the current sample with tag set $\mathcal{C}\subseteq \mathcal{V}_{\mathrm{tag}}$, the schema target is
\begin{equation}
\setlength{\abovedisplayskip}{3pt}
\setlength{\belowdisplayskip}{3pt}
\setlength{\abovedisplayshortskip}{2pt}
\setlength{\belowdisplayshortskip}{2pt}
\mathbf{s}
=
\sum_{c\in\mathcal{C}}
W_{\mathrm{tag}}[c]
\in\mathbb{R}^{D}.
\end{equation}
A lightweight prediction head maps the routing signature into the schema embedding space:
\( 
\hat{\mathbf{s}}\!=\!H_{\mathrm{pred}}(\mathbf{g}).
\)
The SAPR loss encourages samples with different evidence structures to induce distinguishable global routing signatures:
\begin{equation}  
\setlength{\abovedisplayskip}{3pt}
\setlength{\belowdisplayskip}{3pt}
\setlength{\abovedisplayshortskip}{2pt}
\setlength{\belowdisplayshortskip}{2pt}
\mathcal{L}_{\mathrm{SAPR}}
=
1-
\frac{
\hat{\mathbf{s}}^{\top}\mathbf{s}
}{
\|\hat{\mathbf{s}}\|_2\|\mathbf{s}\|_2
}.
\end{equation}
The final supervised fine-tuning objective is
\begin{equation} 
\setlength{\abovedisplayskip}{3pt}
\setlength{\belowdisplayskip}{3pt}
\setlength{\abovedisplayshortskip}{2pt}
\setlength{\belowdisplayshortskip}{2pt}
\mathcal{L}_{\mathrm{SFT}}
=
\mathcal{L}_{\mathrm{gen}}
+
\lambda_{\mathrm{lb}}\mathcal{L}_{\mathrm{lb}}
+
\lambda_{\mathrm{SAPR}}\mathcal{L}_{\mathrm{SAPR}},
\end{equation}
where $\mathcal{L}_{\mathrm{gen}}$ denotes the autoregressive negative log-likelihood on the target reasoning and answer, and $\mathcal{L}_{\mathrm{lb}}$ denotes the standard MoE load-balancing loss.

\subsection{Route-Aware MoE Reinforcement Learning (RMRL)}
\label{sec:rl}

Schema-aligned supervised fine-tuning encourages routing signatures to encode the evidence structure of each sample, but the model is still trained primarily by matching target outputs. A correct response does not necessarily indicate that the model used the appropriate modality, resolved the relevant audio-visual relation, or focused on the decisive temporal region. We therefore introduce a route-aware reinforcement learning stage that assigns a reward to complete rollouts and propagates this feedback to both token generation and expert routing.
For each input $x$, the old policy $\pi_{\theta_{\mathrm{old}}}$ generates a group of $G$ rollouts
\(  
\{(o^{(i)}, \mathcal{E}^{(i)})\}_{i=1}^{G},
\)
where $o^{(i)}$ is the generated output sequence and $\mathcal{E}^{(i)}$ denotes the collection of routing decisions over generated tokens and MoE layers. Each rollout receives a composite reward
\begin{equation}
\setlength{\abovedisplayskip}{3pt}
\setlength{\belowdisplayskip}{3pt}
\setlength{\abovedisplayshortskip}{2pt}
\setlength{\belowdisplayshortskip}{2pt}
\mathcal{R}(x,o^{(i)})
=
\lambda_{\mathrm{ans}}\mathcal{R}_{\mathrm{ans}}(x,o^{(i)})
+
\lambda_{\mathrm{mcr}}\mathcal{R}_{\mathrm{mcr}}(x,o^{(i)})
+
\lambda_{\mathrm{ctg}}\mathcal{R}_{\mathrm{ctg}}(x,o^{(i)}).
\end{equation}
Here, $\mathcal{R}_{\mathrm{ans}}$ is a binary answer-correctness reward computed from the normalized final answer. $\mathcal{R}_{\mathrm{mcr}}$ is a Modality-Consistent Reasoning reward that checks whether the generated reasoning attributes the answer to the appropriate visual and/or audio evidence and captures their relation when needed. We gate $\mathcal{R}_{\mathrm{mcr}}$ by $\mathcal{R}_{\mathrm{ans}}$ to avoid rewarding fluent but incorrect rationales. $\mathcal{R}_{\mathrm{ctg}}$ is the Cognitive Temporal Grounding reward defined in \cref{sec:ctg}, which measures whether the model's final reasoning state focuses on the annotated evidence span. Full reward definitions and verifier prompts are provided in Appendix~\ref{app:reward_details}. 
We convert group rewards into a relative advantage
\begin{equation}
\setlength{\abovedisplayskip}{3pt}
\setlength{\belowdisplayskip}{3pt}
\setlength{\abovedisplayshortskip}{2pt}
\setlength{\belowdisplayshortskip}{2pt}
\hat{A}^{(i)}
=
\frac{
\mathcal{R}(x,o^{(i)})
-
\operatorname{mean}\left(\{\mathcal{R}(x,o^{(j)})\}_{j=1}^{G}\right)
}{
\operatorname{std}\left(\{\mathcal{R}(x,o^{(j)})\}_{j=1}^{G}\right)
+
\epsilon_A
}.
\end{equation}

For generated token $t$ in rollout $i$, the token branch follows the clipped GRPO surrogate:
\begin{equation}
\setlength{\abovedisplayskip}{3pt}
\setlength{\belowdisplayskip}{3pt}
\setlength{\abovedisplayshortskip}{2pt}
\setlength{\belowdisplayshortskip}{2pt}
\resizebox{0.9\linewidth}{!}{$\displaystyle
\mathcal{J}_{\mathrm{token}}
=
-\frac{1}{G}
\sum_{i=1}^{G}
\sum_{t}
\left(
\frac{
\pi_{\theta}\!\left(o^{(i)}_{t}\mid x,o^{(i)}_{<t},\mathcal{E}^{(i)}_{<t}\right)
}{
\pi_{\theta_{\mathrm{old}}}\!\left(o^{(i)}_{t}\mid x,o^{(i)}_{<t},\mathcal{E}^{(i)}_{<t}\right)
}
\hat{A}^{(i)},
\operatorname{clip}\!\left(
\frac{
\pi_{\theta}\!\left(o^{(i)}_{t}\mid x,o^{(i)}_{<t},\mathcal{E}^{(i)}_{<t}\right)
}{
\pi_{\theta_{\mathrm{old}}}\!\left(o^{(i)}_{t}\mid x,o^{(i)}_{<t},\mathcal{E}^{(i)}_{<t}\right)
},
1-\epsilon,1+\epsilon
\right)
\hat{A}^{(i)}
\right)
$}
\end{equation}

To propagate rollout-level feedback to expert allocation, we define an analogous gate branch. Let $\mathcal{E}^{(i)}_{t,l}$ denote the selected top-$k$ expert set for generated token $t$ at MoE layer $l$ in rollout $i$. Since the hard top-$k$ selection is non-differentiable, we score the selected expert set using the dense pre-top-$k$ router distribution. Let
\( 
\mathbf{p}^{(i)}_{t,l}(\theta)
=
\rho_{\theta}^{l}(\mathbf{h}^{(i)}_{t,l})
\in \Delta^{N-1}
\)
be the router distribution under the current policy. We define the likelihood surrogate for the selected expert set as
\begin{equation}
\setlength{\abovedisplayskip}{3pt}
\setlength{\belowdisplayskip}{3pt}
\setlength{\abovedisplayshortskip}{2pt}
\setlength{\belowdisplayshortskip}{2pt}
\ell_{\theta}^{l}
\left(
\mathcal{E}^{(i)}_{t,l}
\mid
\mathbf{h}^{(i)}_{t,l}
\right)
=
\prod_{e\in \mathcal{E}^{(i)}_{t,l}}
\mathbf{p}^{(i)}_{t,l}(\theta)[e].
\end{equation}
The old router provides the corresponding behavior likelihood, yielding the gate objective
\begin{equation}
\resizebox{0.9\linewidth}{!}{$\displaystyle
\mathcal{J}_{\mathrm{gate}}
=
-\frac{1}{G}
\sum_{i=1}^{G}
\sum_{t,l}
\left(
\frac{
\ell_{\theta}^{l}
\left(
\mathcal{E}^{(i)}_{t,l}
\mid
\mathbf{h}^{(i)}_{t,l}
\right)
}{
\ell_{\theta_{\mathrm{old}}}^{l}
\left(
\mathcal{E}^{(i)}_{t,l}
\mid
\mathbf{h}^{(i)}_{t,l}
\right)
}
\hat{A}^{(i)},
\operatorname{clip}
\left(
\frac{
\ell_{\theta}^{l}
\left(
\mathcal{E}^{(i)}_{t,l}
\mid
\mathbf{h}^{(i)}_{t,l}
\right)
}{
\ell_{\theta_{\mathrm{old}}}^{l}
\left(
\mathcal{E}^{(i)}_{t,l}
\mid
\mathbf{h}^{(i)}_{t,l}
\right)
},
1-\epsilon,
1+\epsilon
\right)
\hat{A}^{(i)}
\right).
$}
\end{equation}

Both branches use the same rollout-level advantage $\hat A^i$, so a high-reward rollout increases the likelihood of both the generated tokens and the routing decisions that produced them. We add KL regularization to the frozen reference model for both the language policy and router, and keep the standard load-balancing loss for the gate branch. The overall reinforcement objective is
\[
\min_{\theta}
\mathcal{J}_{\mathrm{RL}}
=
\mathcal{J}_{\mathrm{token}}
+
\eta\mathcal{J}_{\mathrm{gate}}
+
\beta_{\mathrm{tok}}\mathrm{KL}_{\mathrm{tok}}
+
\beta_{\mathrm{gate}}\mathrm{KL}_{\mathrm{gate}}
+
\lambda_{\mathrm{lb}}\mathcal{L}_{\mathrm{lb}}.
\]
This objective lets outcome-level reward update both the generated sequence and the route decisions that produced it. The token branch improves the generated reasoning and final answer, while the gate branch directly updates expert allocation using feedback from complete rollouts.

\subsection{Cognitive Temporal Grounding Reward}
\label{sec:ctg}
Answer-level reward can assign high scores to responses that predict the correct answer without using the short event window that contains the relevant social cue. We therefore introduce a reward that encourages the model's final reasoning state to focus on the annotated evidence span. We use the model's own attention over audio-visual tokens as a lightweight process signal as a differentiable proxy for temporally grounded reasoning.\looseness-1
 
The omni backbone aligns audio and video tokens on a shared temporal grid through time-aware position encoding. Let $\mathcal{I}_{\mathrm{av}}$ denote the set of audio and video input-token indices, and let $\tau_j$ be the temporal index associated with token $j$. Rather than probing a single generated token, we use a short suffix of the reasoning span, denoted $\mathcal{Q}_{\mathrm{suffix}}$, including the closing reasoning token. For each audio-visual input token $j$, we average the native causal attention from these query positions over a set of deep layers $\mathcal{L}_{\mathrm{ctg}}$ and attention heads, which gives a token-level attention summary $\bar a(j)$ over the audiovisual history
\begin{equation}
\setlength{\abovedisplayskip}{3pt}
\setlength{\belowdisplayskip}{3pt}
\setlength{\abovedisplayshortskip}{2pt}
\setlength{\belowdisplayshortskip}{2pt}
\bar{a}(j)
=
\frac{1}{|\mathcal{Q}_{\mathrm{suffix}}|}
\frac{1}{|\mathcal{L}_{\mathrm{ctg}}|}
\sum_{u\in\mathcal{Q}_{\mathrm{suffix}}}
\sum_{l\in\mathcal{L}_{\mathrm{ctg}}}
\frac{1}{H_l}
\sum_{h=1}^{H_l}
A^{l,h}_{u\rightarrow j},
\end{equation}
where $A^{l,h}_{u\rightarrow j}$ is the causal attention weight from query position $u$ to input token $j$ at layer $l$ and head $h$.
Since multiple audio and video tokens can share the same temporal index, we pool token-level attention into a distribution over time bins:
\begin{equation}
\setlength{\abovedisplayskip}{3pt}
\setlength{\belowdisplayskip}{3pt}
\setlength{\abovedisplayshortskip}{2pt}
\setlength{\belowdisplayshortskip}{2pt}
\alpha(\tau)
=
\frac{1}{Z_{\alpha}}
\sum_{j\in\mathcal{I}_{\mathrm{av}}:\tau_j=\tau}
\bar{a}(j),
\end{equation}
where $Z_{\alpha}$ normalizes $\alpha$ to sum to one. This converts attention over raw audio-visual tokens into a temporal focus distribution over the physical timeline of the clip. Because the evidence annotations are unified time spans, audio and video tokens are pooled together at this stage.
Let
\( 
\mathcal{S}_{\mathrm{gt}}
=
\{[\tau^{s}_{u},\tau^{e}_{u}]\}_{u=1}^{M}
\)
be the annotated evidence spans for the sample after mapping them onto the same temporal grid. We convert these intervals into a smooth target distribution using a Gaussian proximity kernel:
\begin{equation}
\setlength{\abovedisplayskip}{3pt}
\setlength{\belowdisplayskip}{3pt}
\setlength{\abovedisplayshortskip}{2pt}
\setlength{\belowdisplayshortskip}{2pt}
P_{\mathrm{gt}}(\tau)
=
\frac{1}{Z_{\mathrm{gt}}}
\max_{[\tau^{s},\tau^{e}]\in\mathcal{S}_{\mathrm{gt}}}
\exp
\left(
-\frac{
d(\tau,[\tau^{s},\tau^{e}])^2
}{
2\sigma^2
}
\right),
\end{equation}
where $d(\tau,[\tau^{s},\tau^{e}])=0$ if $\tau$ lies inside the interval and otherwise equals the distance to the nearest interval boundary, and 
 $Z_{gt}$ normalizing constant. This smooth target gives tolerance near span boundaries and avoids forcing all probability mass onto a single timestamp. For numerical stability, we smooth and renormalize the model's temporal focus distribution:
\begin{equation}
\setlength{\abovedisplayskip}{3pt}
\setlength{\belowdisplayskip}{3pt}
\setlength{\abovedisplayshortskip}{2pt}
\setlength{\belowdisplayshortskip}{2pt}
\tilde{\alpha}(\tau)
=
\frac{
\alpha(\tau)+\epsilon_{\mathrm{ctg}}
}{
\sum_{\tau'}\left(\alpha(\tau')+\epsilon_{\mathrm{ctg}}\right)
}.
\end{equation}
The Cognitive Temporal Grounding reward is then defined as
\begin{equation}
\mathcal{R}_{\mathrm{ctg}}(x,o)
=
\begin{cases}
\exp\!\left(
-\gamma
D_{\mathrm{KL}}
\left(
P_{\mathrm{gt}}
\,\|\, 
\tilde{\alpha}
\right)
\right),
& \text{if a valid reasoning suffix exists},\\
0,
& \text{otherwise}.
\end{cases}
\end{equation}
where $\tilde{\alpha}(\tau) \propto \alpha(\tau) + \epsilon_{\mathrm{ctg}}$ is an $\epsilon$-smoothed version of $\alpha$ for numerical stability. Because $P_{gt}$ and $\tilde{\alpha}$ are both normalized distributions over the same temporal grid, the KL divergence measures how well the model's temporal focus matches the annotated evidence region. 
A high reward is obtained when the model's final reasoning state places attention mass near the annotated evidence span. This reward uses the model's own attention and does not require a separate temporal grounding model. As a result, it gives process-level feedback in cases where answer correctness alone cannot distinguish genuine use of the relevant cue from shortcut-based prediction.

\begin{figure}[!t]
    \centering    
    \includegraphics[width=\linewidth]{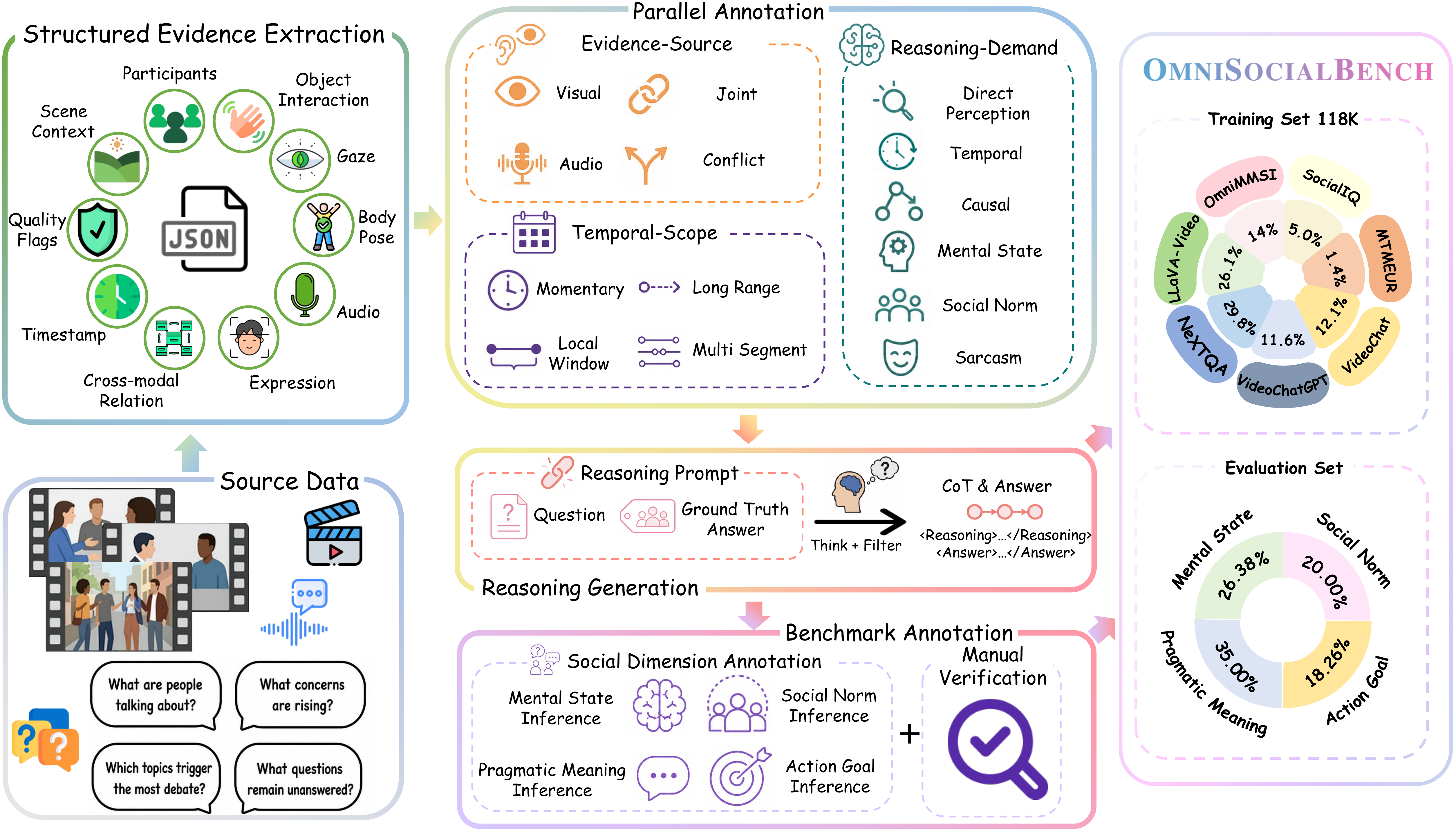}
    \caption{\textbf{\benchmarknamegradient{} Dataset and Annotation Pipeline.} \benchmarknamenc{} augments video QA examples with structured audio-visual evidence, schema labels, grounded reasoning traces, and temporal evidence spans. The pipeline extracts observable social cues, assigns cross-modal relations, reasoning demands, and temporal-scope annotations, and generates evidence-grounded rationales. The evaluation split is manually verified for diagnostic social omni reasoning.}
    \label{fig:dataset}
\end{figure}

\section{\benchmarknamegradient{} Diagnostic Benchmark for Social Omni Reasoning}
\label{sec:data}
Existing audio-visual QA datasets typically supervise only final answers, leaving the supporting
evidence structure implicit. This is limiting for social video reasoning, where the answer may depend
on facial expression, gesture, speech content, vocal tone, temporal ordering, or a mismatch between
what is said and what is visually expressed. We therefore construct \benchmarknamenc{} as a
diagnostic social video QA resource that exposes not just the answer, but also the modality, reasoning, and temporal structure required to support it.\looseness-1

\benchmarknamenc{} contains a structured training split with 118K examples and a manually verified evaluation split for diagnostic benchmarking. We build \benchmarknamenc{} from diverse sources covering multi-person interaction, affective and pragmatic reasoning, egocentric scenarios, and general video understanding requiring social interpretation. Each example preserves the original video clip, audio track, question, and ground-truth answer, and augments them with structured audio-visual evidence, schema labels, grounded reasoning, and temporal evidence spans. The evidence records observable cues such as participants, scene context, gaze, expression, body pose, gesture,
object interaction, speech content, speaker changes, laughter, silence, and tone-relevant acoustic events. The reasoning trace explains the answer using this recorded evidence, while the temporal spans identify the time region supporting the answer. Figure~\ref{fig:dataset} summarizes the dataset.

\noindent \textbf{Schema tags.}
The schema tag set defines which modality matters, what reasoning is required, and where the
supporting cue lies in time. Each example receives one tag from each of three axes. The
\texttt{evidence\_source} axis specifies whether the answer is primarily
\textsc{Visual}, \textsc{Audio}, \textsc{Joint}, 
\textsc{Conflict}. The \texttt{reasoning\_demand} axis specifies the required inference
operation: \textsc{Direct Perception}, \textsc{Temporal}, \textsc{Causal},
\textsc{Mental State}, \textsc{Social Norm}, or
\textsc{Sarcasm}. The \texttt{temporal\_scope} axis specifies whether the evidence
is a \textsc{Momentary}, \textsc{Local Window}, \textsc{Long Range}, or
\textsc{Multi Segment}.

\noindent \textbf{Grounded supervision and evaluation.}
These annotations directly support \modelnamenc{}: schema labels supervise Schema-Aligned
Predictive Routing, grounded reasoning supports supervised fine-tuning and modality-consistent rewards, and temporal spans supervise the Cognitive Temporal Grounding reward. The training split provides supervision for schema-guided routing and reward learning, while the evaluation split measures whether models answer social questions using the appropriate modality, reasoning operation, and temporal evidence. The evaluation split also includes a \texttt{social\_dimension} label for analysis, covering \textsc{Mental State Inference}, \textsc{Pragmatic Meaning Inference},
\textsc{Action Goal Inference}, and \textsc{Social Norm Inference}. All evaluation examples are manually verified for answer validity, evidence consistency, schema correctness, reasoning support, temporal grounding, and social-dimension assignment. Details on data sources, filtering, and verification are provided in Appendix~\ref{app:annotation_protocol}, while  \benchmarknamenc{} examples covering diverse social reasoning cases and multimodal evidence cues are available in Appendix~\ref{app:benchexample}.

\section{Experiments}
We evaluate \modelnamegradient{} on the OmniSocial Benchmark as well as a diverse suite of public audio-visual, audio-only, and video-only benchmarks to comprehensively assess omni-modal perception and reasoning capabilities.
Training implementation details are provided in Appendix~\ref{sec:training_detail}.

\begin{table}[t]
\centering
\caption{\textbf{Social understanding results across categories.}
\colorbox{color1}{\phantom{\rule{1ex}{1ex}}} denotes VLMs that do not support omni inputs.
\colorbox{color3}{\phantom{\rule{1ex}{1ex}}} denotes open-source omni models.
\colorbox{color2}{\phantom{\rule{1ex}{1ex}}} denotes VLMs that take omni inputs.
\colorbox{color4}{\phantom{\rule{1ex}{1ex}}} denotes our \modelnamegradient{} model.
Numbers represent accuracy ($\uparrow$). \textbf{Bold} indicates the best result in each column.}
\resizebox{0.99\textwidth}{!}{%
\begin{tabular}{lccccc}
\toprule
 & \textbf{Social Norm} & \textbf{Action Goal} & \textbf{Mental State} & \textbf{Pragmatic Meaning} & \textbf{Avg}\\
\midrule
\rowcolor{color1} GPT 5.4~\cite{singh2025openai}    
& 14.37 & 20.13 & 18.96 & 18.57 & 18.12 \\
\rowcolor{color1} Qwen 3.6 Plus~\cite{yang2025qwen3}   
& 15.00 & 26.17 & 14.69 & 19.29 & 18.50 \\
\rowcolor{color1} GLM 5V Turbo~\cite{hong2025glm}  
& 14.37 & 18.12 & 13.27 & 17.50 & 15.88 \\
\midrule
\rowcolor{color3} Baichuan-Omni-1.5~\cite{li2025baichuan}     
& 14.37 & 12.75 & 12.80 & 12.86 & 13.12 \\
\rowcolor{color3} OmniVinci~\cite{ye2025omnivinci}            
& 18.75 & 20.81 & 26.07 & 21.07 & 21.88 \\
\rowcolor{color3} Qwen3 Omni~\cite{xu2025qwen3}        
& 26.11 & 33.02 & 38.97 & 31.44 & 32.61 \\
\midrule
\rowcolor{color2} MiMo V2 Omni~\cite{xiao2026mimo}           
& 30.63 & 32.89 & 31.28 & 30.36 & 31.13 \\
\rowcolor{color2} Qwen 3.5 Omni~\cite{team2026qwen3}       
& 31.25 & 30.87 & 36.02 & 28.57 & 31.50 \\
\rowcolor{color2} Gemini 3.1 Flash~\cite{comanici2025gemini}          
& 27.50 & 47.65 & 43.33 & 41.79 & 40.43 \\
\rowcolor{color2} Gemini 3.1 Pro~\cite{comanici2025gemini}           
& 33.33 & 36.84 & 60.00 & 43.90 & 44.05 \\
\midrule
\rowcolor{color4} \modelnamegradient{} \textbf{(Ours)} 
& \textbf{58.13} & \textbf{66.44} & \textbf{60.18} & \textbf{55.71} & \textbf{59.38} \\
\bottomrule
\end{tabular}%
}
\label{tab:benchmark_results}
\vspace{-0.3cm}
\end{table}

\textbf{\benchmarknamenc{} Social Understanding Results.}
Table~\ref{tab:benchmark_results} reports results on the OmniSocial Benchmark across four social inference categories. Among open-source omni models, Qwen3 Omni achieves the strongest performance with an average of 32.61\%, while VLMs that take omni inputs reach higher scores, led by Gemini 3.1 Pro at 44.05\%. \modelnamegradient{} achieves the best result in every category, with an average of 59.38, surpassing Gemini 3.1 Pro by over 15\% and more than doubling the performance of most open-source omni baselines. The largest gains appear on Action Goal (+18.79\% over the strongest baseline) and Social Norm (+24.80\%) inference, where joint reasoning over visual and audio cues is most critical. These results indicate that schema-guided routing and route-aware reinforcement learning yield substantial improvements on social reasoning tasks that require coordinated use of visual and audio evidence. We provide some examples in Appendix~\ref{app:response}, where we compare the response of \modelnamenc{} and Gemini 3.1 Pro. 

\textbf{Public Audio-Visual Benchmarks.} We compare \modelnamenc{} against Qwen3-Omni-30B-A3B~\citep{xu2025qwen3} on ten benchmarks spanning audio-visual, audio-only, and video-only understanding. For joint audio-visual reasoning, we evaluate on \textsc{OmniBench}~\citep{li2024omnibench}, \textsc{WorldSense}~\citep{hong2025worldsense},  \textsc{Daily-Omni}~\citep{zhou2025daily}, AV- \textsc{SpeakerBench}~\citep{nguyen2025see},  \textsc{AV-Odyssey}~\citep{gong2024av}, and  \textsc{OmniVideoBench}~\citep{li2025omnivideobench}, which probe cross-modal integration across varied temporal scales and conditions. For audio-only reasoning, we use  \textsc{MMAU}~\citep{sakshi2024mmau}, and for video-only understanding we include  \textsc{Video-MME-v2}~\citep{fu2026video},  \textsc{MMVU}~\citep{zhao2025mmvu}, and  \textsc{MVBench}~\citep{li2024mvbench}.
As shown in Table~\ref{tab:av_benchmark},  \modelnamenc{} improves over Qwen3-Omni-30B-A3B on most audio-visual benchmarks, indicating that schema-guided routing enhances the model's ability to select and integrate the appropriate modalities.
The small performance drop on audio-centric evaluations is consistent with reallocating expert capacity toward joint audio-visual coordination. Consistent gains on video-only benchmarks further suggest that the cognitive schema benefits temporal and reasoning-heavy tasks even when audio is less central. 
These findings support that \modelnamenc{} delivering consistent gains over strong baselines, improving the underlying modality-aware reasoning process, and that these benefits generalize beyond the proposed OmniSocial benchmark.

\begin{table}[t]
\centering
\caption{\textbf{Audio-visual benchmark comparison between Qwen3-Omni-30B and \modelnamegradient{} across 10 benchmarks.} Numbers represent accuracy ($\uparrow$). \textbf{Bold} indicates the best in each column.}
\label{tab:av_benchmark}

\resizebox{\textwidth}{!}{%
\begin{tabular}{lccccc}
\toprule
& \textbf{AV-\textsc{SpeakerBench}} & \textbf{\textsc{OmniBench}} & \textbf{\textsc{MMVU}} & \textbf{\textsc{WorldSense}} & \textbf{\textsc{Daily-Omni}} \\
\midrule
Qwen3 Omni 30B~\cite{xu2025qwen3}
& 54.1 & 55.5 & 59.8 & \textbf{54.0} & 73.6 \\
\rowcolor{color4} \modelnamegradient{} \textbf{(Ours)}
& \textbf{56.2} & \textbf{56.7} & \textbf{60.7} & 53.8 & \textbf{74.3} \\
\bottomrule
\end{tabular}%
}

\vspace{0.5em}

\resizebox{\textwidth}{!}{%
\begin{tabular}{lccccc}
\toprule
& \textbf{\textsc{AV-Odyssey}} & \textbf{\textsc{MVBench}} & \textbf{\textsc{MMAU}} & \textbf{\textsc{OmniVideoBench}} & \textbf{\textsc{Video-MME-v2}} \\
\midrule
Qwen3 Omni 30B~\cite{xu2025qwen3}
& 34.7 & 69.5 & \textbf{75.4} & 38.4 & 36.5 \\
\rowcolor{color4} \modelnamegradient{} \textbf{(Ours)}
& \textbf{36.8} & \textbf{70.2} & 73.2 & \textbf{38.9} & \textbf{36.8} \\
\bottomrule
\end{tabular}%
}
\end{table}

\begin{figure}[t]
    \centering    
    \includegraphics[width=\linewidth]{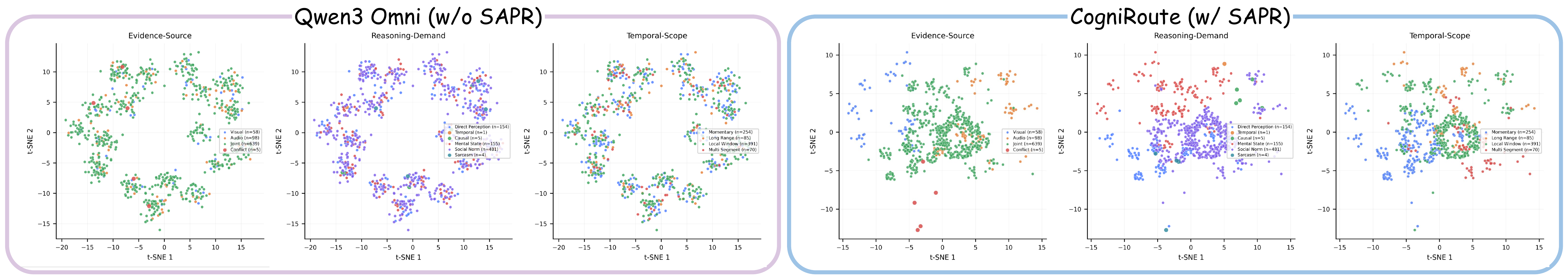}
    \caption{\textbf{Routing signature visualization.} We visualize the MoE routing signatures of benchmark samples before and after applying SAPR. For each model, the same two-dimensional coordinates are colored by Cross Modal Relation, Reasoning Demand, and Temporal Scope. }
    \label{fig:routing_tsne}
\end{figure}

\textbf{Routing signature visualization.}
Figure~\ref{fig:routing_tsne} visualizes the MoE routing signatures before and after applying SAPR. Without SAPR, routing assignments are diffuse and weakly structured, with substantial overlap across samples. After SAPR, routing signatures form more distinct and coherent clusters that align with \texttt{evidence\_source}, \texttt{reasoning\_demand}, and \texttt{temporal\_scope}. This indicates that SAPR reduces routing noise and induces more structured routing patterns, where samples with similar evidence structure are more likely to share expert usage. This suggests that routing becomes more interpretable and better aligned with the reasoning requirements of the task.

\textbf{Ablations.}
Figure~\ref{fig:core_ablation_avg} summarizes the main ablations, with full per-category results provided in the appendix. SAPR improves standard SFT from 40.13 to 50.13 average accuracy, while corrupted schema supervision yields 
\begin{wrapfigure}{l}{0.7\textwidth}
\vspace{-0.45cm}
\centering
\begin{tikzpicture}

\begin{axis}[
    name=saprplot,
    ybar,
    width=0.39\textwidth,
    height=3.65cm,
    ymin=38,
    ymax=52,
    title={\scriptsize SAPR},
    ylabel={Avg. Acc. (\%)},
    symbolic x coords={SFT, Rand., Shuf., Shared, Full},
    xtick=data,
    xticklabels={SFT,Rand.,Shuf.,Shared,Full},
    xticklabel style={font=\tiny, rotate=30, anchor=center},
    yticklabel style={font=\tiny},
    ylabel style={font=\tiny},
    title style={font=\scriptsize, yshift=-1mm},
    axis x line*=bottom,
    axis y line*=left,
    axis line style={gray!55},
    tick style={draw=none},
    ymajorgrids=true,
    grid style={gray!18},
    bar width=8pt,
    enlarge x limits=0.13,
    nodes near coords,
    nodes near coords align={vertical},
    every node near coord/.append style={
        font=\tiny,
        yshift=1pt,
        /pgf/number format/fixed,
        /pgf/number format/precision=1
    },
]
\addplot[fill=saprE, draw=none] coordinates {
    (SFT,40.13)
    (Rand.,40.63)
    (Shuf.,41.38)
    (Shared,44.50)
    (Full,50.13)
};
\end{axis}

\begin{axis}[
    at={(saprplot.outer east)},
    anchor=outer west,
    xshift=0.12cm,
    ybar,
    width=0.39\textwidth,
    height=3.65cm,
    ymin=48,
    ymax=61,
    title={\scriptsize RMRL},
    symbolic x coords={SFT, Tok., Tok.+R, Gate, Full},
    xtick=data,
    xticklabels={SFT,Tok.,Tok.+R,Gate,Full},
    xticklabel style={font=\tiny, rotate=30, anchor=center},
    yticklabel style={font=\tiny},
    title style={font=\scriptsize, yshift=-1mm},
    axis x line*=bottom,
    axis y line*=left,
    axis line style={gray!55},
    tick style={draw=none},
    ymajorgrids=true,
    grid style={gray!18},
    bar width=8pt,
    enlarge x limits=0.13,
    nodes near coords,
    nodes near coords align={vertical},
    every node near coord/.append style={
        font=\tiny,
        yshift=1pt,
        /pgf/number format/fixed,
        /pgf/number format/precision=1
    },
]
\addplot[fill=rlE!75, draw=none] coordinates {
    (SFT,50.13)
    (Tok.,56.13)
    (Tok.+R,56.75)
    (Gate,51.88)
    (Full,59.38)
};
\end{axis}
\end{tikzpicture}
\caption{\textbf{Core component ablations.} Average accuracy for SAPR design and RMRL token/gate optimization.}
\label{fig:core_ablation_avg}
\end{wrapfigure}
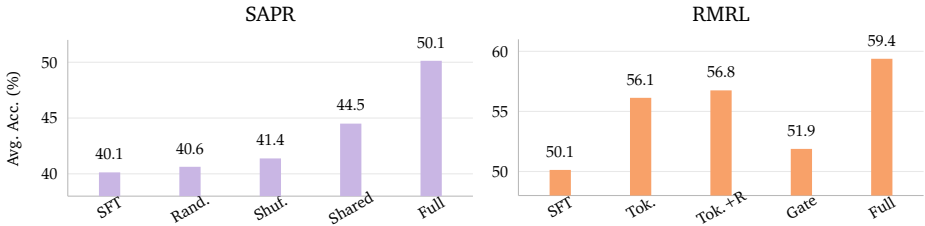only marginal gains, showing that the benefit comes from aligning routing with the correct evidence structure. 
Starting from the same SAPR-trained checkpoint, routing-aware RL (RMRL) further improves performance to 59.38. Token-level RL provides strong gains, but the full token-and-gate objective performs best, indicating that explicitly optimizing expert allocation contributes beyond token generation alone. Full per-category results and additional ablation studies on SAPR design, schema supervision quality, tag-embedding collapse, token-versus-gate optimization, and routing behavior under gate optimization are provided in Appendix~\ref{sec:ablationsmore}.

\begin{figure}[t]
    \centering
    \includegraphics[width=\linewidth]{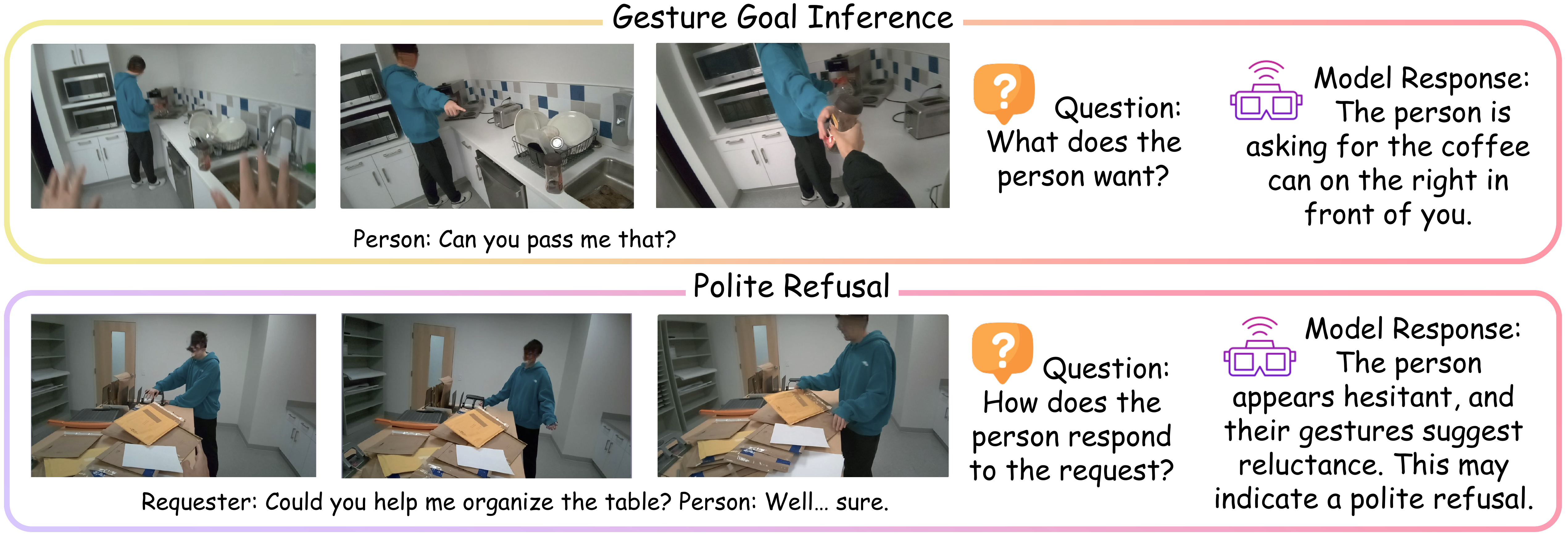}
    \caption{\textbf{Real-world VR Application} illustrating \modelnamenc{}'s ability to infer human intent.}
    \label{fig:VR}
\end{figure}
\textbf{Real-world VR Smart Glasses Deployment.} We deploy \modelnamenc{} on VR glasses for real-time social interaction, as shown in Figure~\ref{fig:VR}. In the first example (top row), \modelnamenc{} performs synergistic routing by combining hand trajectories with partial speech to provide contextual assistance. In the second example (bottom row), \modelnamenc{} captures subtle social cues, detecting hesitation and conflicting gestures, to infer true intent (\eg polite refusal).

\section{Conclusion}
We introduce \modelnamenc{}, a schema-guided MoE framework for omni-modal social reasoning that aligns expert routing with the evidence structure of each question. By combining Schema-Aligned Predictive Routing with route-aware reinforcement learning, \modelnamenc{} improves accuracy while encouraging more evidence-aware computation across modalities and time. Across our \benchmarknamenc{} benchmark and public audio-visual tasks, \modelnamenc{} shows consistent gains over strong omni baselines, especially on questions requiring cross-modal coordination, conflict resolution, and temporally grounded social inference. Results show that evidence-structured routing is an effective training signal for improving social omni-modal reasoning.\looseness-1

\bibliographystyle{plainnat}
\bibliography{main}

\newpage
\appendix

\section{Reward Details}
\label{app:reward_details}

\subsection{Answer Correctness Reward}

We use a binary answer-correctness reward. After extracting the final answer from the model output, we compare it against the ground-truth answer under the benchmark evaluation protocol. A correct answer receives a score of 1, and an incorrect answer receives a score of 0. This reward is deterministic and does not rely on an external judge model.

\subsection{Modality-Consistent Reasoning Reward}

\begin{figure}[ht!]
    \centering    
    \includegraphics[width=\linewidth]{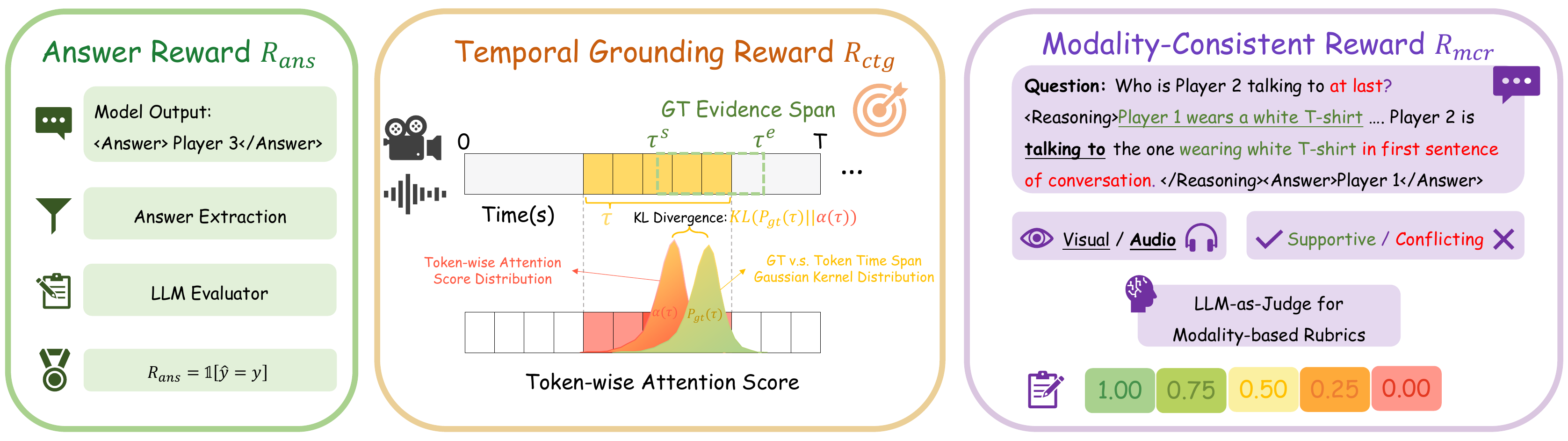}
    \caption{\textbf{Visualization of the reward design.} The overall reward consists of three complementary components: the answer reward $\mathcal{R}_{\mathrm{ans}}$ encourages correct final answers, the Cognitive Temporal Grounding reward $\mathcal{R}_{\mathrm{ctg}}$ guides the model to attend to the annotated temporal evidence span, and the Modality-Consistent Reasoning reward $\mathcal{R}_{\mathrm{mcr}}$ promotes reasoning grounded in the required visual and/or audio evidence.}
    \label{fig:reward}
\end{figure}

While the Cognitive Temporal Grounding reward encourages the model to focus on the correct temporal region, it does not directly assess whether the generated reasoning grounds the answer in the appropriate modality evidence. To complement it, we introduce a Modality-Consistent Reasoning reward computed by a frozen LLM judge (\eg Gemini-3.1-Pro).
This reward is built upon the annotations produced by our Structured Evidence Extraction pipeline. For each sample, the annotation specifies whether the reasoning is expected to rely on visual evidence, audio evidence, or both, and, for dual-modality cases, whether the relation between the two modalities is supportive or conflicting.
During reinforcement learning, we extract the text span between \texttt{<Reasoning>} and \texttt{</Reasoning>} and ask the frozen judge to score how well the reasoning matches the target modality requirement. The judge outputs a scalar score in $[0,1]$ using a five-level rubric: \{0, 0.25, 0.5, 0.75, 1.0\}.
A score of {1.0} is assigned when the reasoning explicitly uses the required modality evidence with concrete observations and clearly links them to the answer. When both modalities are required, the reasoning must also correctly characterize their relation, \eg whether they are mutually supportive or in conflict. A score of {0.75} indicates that the reasoning is largely modality-consistent but remains incomplete, weakly connected to the answer, or only partially captures the cross-modal relation. A score of {0.5} is used when the reasoning only partially satisfies the requirement, such as relying on only one required modality or mentioning the correct modality without concrete evidence. A score of {0.25} is assigned when the reasoning contains weak, vague, or ambiguous modality cues or mixes correct and incorrect evidence sources. A score of {0} is assigned when the reasoning fails to ground the answer in the required modality, relies on the wrong evidence source, or does not contain a valid reasoning span.
Importantly, the judge is instructed to reward concrete modality-specific observations rather than superficial keyword matching. For example, statements such as ``the speaker sounds hesitant'' or ``the person smiles while stepping back'' are treated as grounded evidence, whereas generic phrases such as ``based on the video'' or ``from the audio'' without specific observations are not sufficient for a high score. We keep the judge frozen and use the same prompt throughout all experiments. The full prompt is shown in Fig.~\ref{fig:app_structure_prompt}. The judge evaluates only the modality consistency of the reasoning, while answer correctness is handled separately by the binary answer reward.
We also provide a visualization of the reward design in Fig.~\ref{fig:reward}.

\begin{figure}[ht!]
\centering
\resizebox{1\textwidth}{!}{
\begin{planbox}{Modality-Consistent Reasoning Judge Prompt}
\small

\textbf{System Prompt:}

You are a strict evaluator for modality-consistent reasoning in multimodal question answering.
Given a question, a final answer, a reasoning trace, and a structured evidence annotation, your task is to score whether the reasoning grounds the answer in the required modality evidence.

Evaluate only the reasoning trace.
Do \textbf{not} judge whether the final answer itself is correct.

\textbf{Scoring Rubric:}
\begin{itemize}
    \item \textbf{1.00}: The reasoning explicitly uses the required modality evidence with concrete observations and clearly links them to the answer. If both modalities are required, it also correctly describes their relation (supportive or conflicting).
    \item \textbf{0.75}: The reasoning is mostly modality-consistent, but the evidence grounding is incomplete, weakly connected to the answer, or only partially captures the cross-modal relation.
    \item \textbf{0.50}: The reasoning only partially satisfies the modality requirement, \eg it relies on only one required modality, or mentions the correct modality without concrete evidence.
    \item \textbf{0.25}: The reasoning provides weak, vague, or ambiguous modality evidence, or mixes correct and incorrect evidence sources.
    \item \textbf{0.00}: The reasoning fails to ground the answer in the required modality, relies on the wrong evidence source, or does not contain a usable reasoning span.
\end{itemize}

\textbf{Important Rules:}
\begin{itemize}
    \item Reward \textbf{concrete modality-specific observations}, not superficial keywords.
    \item Generic phrases such as ``based on the video'' or ``from the audio'' do not count as strong evidence unless followed by specific observations.
    \item Visual evidence includes facial expressions, body movements, gaze, scene changes, visible interactions, or other observable events.
    \item Audio evidence includes spoken content, tone, laughter, crying, silence, or salient background sounds.
    \item If both modalities are required, check whether the reasoning uses both and whether it correctly describes their relation.
\end{itemize}

\textbf{Input:}

Question: \{question\}

Final Answer: \{final\_answer\}

Reasoning: \{reasoning\}

Structured Evidence Annotation:
\begin{itemize}
    \item Visual evidence required: \{yes/no\}
    \item Audio evidence required: \{yes/no\}
    \item Cross-modal relation: \{none/support/conflict\}
\end{itemize}

\textbf{Output Format:}

Return a JSON object:
\texttt{\{"score": 0.00/0.25/0.50/0.75/1.00, "brief\_rationale": "..." \}}

\end{planbox}
}
\caption{\textbf{Prompt used for the frozen LLM judge to compute the Modality-Consistent Reasoning reward.} The judge scores whether the generated reasoning grounds the answer in the modality evidence specified by the structured evidence annotation.}
\label{fig:app_structure_prompt}
\end{figure}

\section{Detailed Annotation Protocol}
\label{app:annotation_protocol}

\paragraph{Data Collection and Curation.}
The source pool is drawn from OmniMMSI-YouTube \cite{li2026omni}, SocialIQ \cite{zadeh2019social}, MTMEUR \cite{hu2025beyond}, VideoChat-Conversation \cite{2023videochat}, NeXTQA \cite{xiao2021next}, and LLaVA-Video \cite{zhang2024videoinstructiontuningsynthetic}. Together, these sources cover multi-person interaction, affect-centered scenes, egocentric videos, and general videos that still require social interpretation. A single prompt that tries to observe the clip, solve the question, assign labels, and generate reasoning at the same time is not reliable enough for dataset construction, because it often mixes observation with interpretation and gives an unstable output format. We therefore use a staged pipeline. The model first extracts structured evidence from the clip, and later prompts read the same evidence to assign task labels and generate grounded reasoning. We keep a sample only if the generated answer matches the original ground truth answer after normalization. The evaluation set follows the same pipeline and adds final manual verification.

\subsection{Structured Evidence Extraction}
\label{sec:structure}

In the first stage, the model returns valid JSON only and does not answer the question. The JSON records scene context, participants, timestamped visual events, timestamped audio events, dialogue structure, simple cross modal relations, and quality flags. The extractor is asked to stay at the observation level rather than social interpretation. This gives all later stages a shared evidence source and makes automatic checks simple. 

\subsection{Task Annotation}
\label{sec:annotation}

Each sample, that is, one clip and question pair, receives three labels that depend on the question. \texttt{evidence\_source} describes how visual and audio evidence support the answer, with four classes, \textsc{Visual}, \textsc{Audio}, \textsc{Joint}, and \textsc{Conflict}. \texttt{reasoning\_demand} describes the main reasoning operation, with six classes, \textsc{direct perception}, \textsc{temporal}, \textsc{causal}, \textsc{mental state}, \textsc{social norm}, and \textsc{sarcasm}. \texttt{temporal\_scope} describes how much of the clip must be used, with four classes, \textsc{momentary}, \textsc{local window}, \textsc{long range}, and \textsc{multi segment}. These labels are stored as JSON fields and later guide expert routing over modality, reasoning, and temporal context. After label prediction, the model generates the final target \texttt{<Reasoning>...</Reasoning><Answer>...</Answer>} using only cues already present in the evidence JSON. Samples whose generated answer does not match the original answer after normalization are removed. Detailed prompts and decision rules are given in Appendix~\ref{app:annotation_protocol}.

\subsection{Evaluation Split}
\label{sec:benchmark}

The evaluation split uses the same automatic pipeline, but adds one field used only in the benchmark, \texttt{social\_dimension}, for subset analysis and error reporting. We use four classes, \texttt{Mental State Inference}, \texttt{Pragmatic Meaning Inference}, \texttt{Action Goal Inference}, and \texttt{Social Norm Inference}. This field describes the main social phenomenon tested by the question, while the three task labels above describe the routing path and the required evidence pattern. All benchmark samples receive a final manual verification pass over the evidence JSON, task labels, reasoning text, answer, and \texttt{social\_dimension} before release.

All automatic annotations in this paper were produced with Gemini-3.1-Pro. We use one base model across the full pipeline and change only the task prompt and output constraint. The final target string is \texttt{<Reasoning>...</Reasoning><Answer>...</Answer>}. The fields \texttt{cross\_modal\_relation}, \texttt{reasoning\_demand}, \texttt{temporal\_scope}, and \texttt{social\_dimension} are stored as separate JSON fields, where \texttt{social\_dimension} is used only in \benchmarknamenc{}.

\subsection{Structured Evidence Fields and Extraction Prompt}
\label{app:structure}

A single prompt that asks the model to observe the clip, solve the question, assign task labels, and generate reasoning at the same time often mixes observation with interpretation. We therefore insert a structured evidence step before any label prediction. The structured JSON has seven top level parts: \texttt{scene\_context}, \texttt{participants}, \texttt{visual\_nodes}, \texttt{audio\_nodes}, \texttt{dialogue\_nodes}, \texttt{crossmodal\_nodes}, and \texttt{quality\_flags}. Each node should stay at the observation level. For example, the extractor may record that a speaker looks away, pauses, and speaks with flat prosody, but it should not assign sarcasm or hidden frustration. This shared evidence source makes later labels easier to check.

\begin{figure}[ht!]
\centering
\resizebox{1\textwidth}{!}{
\begin{planbox}{Structured Evidence Extraction Prompt}
\small
\textbf{System.} You are given a pre-segmented video clip with audio, its original question, and its original ground truth answer. Convert the clip into structured evidence JSON. Record only observable cues. Do not answer the question. Do not infer hidden mental states.\\

\textbf{Task.} Use the video and audio to extract scene context, participants, timestamped visual events, timestamped audio events, dialogue structure, simple cross modal relations, and quality flags.\\

\textbf{Required JSON fields.} \texttt{scene\_context}, \texttt{participants}, \texttt{visual\_nodes}, \texttt{audio\_nodes}, \texttt{dialogue\_nodes}, \texttt{crossmodal\_nodes}, \texttt{quality\_flags}.\\

\textbf{Field guidance.} \texttt{scene\_context}: place, social setting, salient objects, camera distance, notable background cues. \texttt{participants}: participant id, role cues, position, visibility, face visibility. \texttt{visual\_nodes}: start time, end time, actor, event type, description. \texttt{audio\_nodes}: start time, end time, speaker, transcript, prosody, non-speech audio. \texttt{dialogue\_nodes}: start time, end time, speaker, addressee, dialogue act, utterance summary. \texttt{crossmodal\_nodes}: start time, end time, relation type, linked visual nodes, linked audio nodes, description. \texttt{quality\_flags}: low light, occlusion, off-screen speaker, weak audio, background noise, uncertainty note.\\

\textbf{Rules.} Every event should have timestamps when possible. Use short factual descriptions. If a field is absent, return an empty list or \texttt{null}. Do not output markdown. Output valid JSON only.
\end{planbox}
}
\caption{\textbf{Structured evidence extraction prompt,} converting each clip into observation-level JSON.}
\label{fig:app_structure_prompt}
\end{figure}

\subsection{Task Labels and Prompts}
\label{app:task_labels}

The three task labels are assigned from the original question, the original ground truth answer, and the structured evidence JSON. They are sample-level labels, so the same clip may receive different labels under different questions.

\paragraph{Evidence Source.}
The field \texttt{evidence\_source} records how the minimum answer evidence is distributed across modalities. \textsc{Visual} is used when the main evidence lies in visual cues such as face, gaze, posture, action, object use, or scene context, and audio adds little useful support. \textsc{Audio} is used when the main evidence lies in transcript, prosody, speaker turn, or other audio cues, while the image is weak or static. \textsc{Joint} is used when the answer needs support from both modalities. \textsc{Conflict} is used when visual and audio cues point to different meanings and the answer depends on resolving the mismatch. This label gives a modality prior to the router.

\begin{figure}[ht!]
\centering
\resizebox{1\textwidth}{!}{
\begin{planbox}{Cross Modal Relation Annotation Prompt}
\small
\textbf{System.} You are given the original question, the original ground truth answer, and the structured evidence JSON for one sample. Assign the single best \texttt{evidence\_source} label. Use the question as the main reference.\\

\textbf{Label definitions.} \textsc{Visual}: the main evidence is in \texttt{visual\_nodes}, \texttt{participants}, or \texttt{scene\_context}. \textsc{Audio}: the main evidence is in \texttt{audio\_nodes} or speaker cues. \textsc{Joint}: the answer needs support from both modalities. \textsc{Conflict}: visual and audio cues point to different meanings, and the answer comes from resolving the mismatch.\\

\textbf{Decision rule.} Choose the label that best matches the minimum evidence needed to support the original answer.\\

\textbf{Output format.} Return valid JSON only in the form \texttt{\{"evidence\_source": "LABEL"\}}.
\end{planbox}
}
\caption{\textbf{Evidence source prompt,} assigning a modality label from the structured evidence JSON.}
\label{fig:app_crossmodal_prompt}
\end{figure}

\paragraph{Reasoning-Demand.}
The field \texttt{reasoning\_demand} records the main reasoning operation required by the question. \textsc{direct perception} is used when the answer follows from directly visible or audible cues. \textsc{temporal} is used when the answer depends on event order, duration, or relation across time. \textsc{causal} is used when the question asks about cause, effect, or trigger. \textsc{mental state} is used for hidden emotion, belief, desire, or intention. \textsc{social norm} is used when the answer depends on role expectation, local rule, or appropriateness in context. \textsc{sarcasm} is used when the surface utterance differs from the intended meaning. This label guides the choice of reasoning experts.

\begin{figure}[ht!]
\centering
\resizebox{1\textwidth}{!}{
\begin{planbox}{Reasoning Demand Annotation Prompt}
\small
\textbf{System.} You are given the original question, the original ground truth answer, and the structured evidence JSON for one sample. Assign the single best \texttt{reasoning\_demand} label. Use the question as the main reference.\\

\textbf{Label definitions.} \textsc{direct perception}: the answer follows from directly observable cues. \textsc{temporal}: the answer needs event order, duration, or relation across time. \textsc{causal}: the answer needs cause, effect, or trigger. \textsc{mental-state}: the answer needs inference about hidden emotion, belief, desire, or intention. \textsc{social-norm}: the answer needs a rule, role expectation, or judgment of appropriateness in context. \textsc{sarcasm}: the intended meaning differs from the literal utterance.\\

\textbf{Decision rule.} Choose the label that best matches the main reasoning step needed to support the original answer.\\

\textbf{Output format.} Return valid JSON only in the form \texttt{\{"reasoning\_demand": "LABEL"\}}.
\end{planbox}
}
\caption{\textbf{Reasoning demand prompt,} assigning a reasoning label from the structured evidence JSON.}
\label{fig:app_reasoning_demand_prompt}
\end{figure}

\paragraph{Temporal-Scope.}
The field \texttt{temporal\_scope} records the smallest temporal field needed to answer the question correctly. \textsc{momentary} is used when a short instant is enough. \textsc{local window} is used when the answer depends on a short continuous span around the key event. \textsc{long range} is used when evidence is spread across a large part of the clip. \textsc{multi segment} is used when several separated spans must be combined. This label guides the temporal field used by the router.

\begin{figure}[ht!]
\centering
\resizebox{1\textwidth}{!}{
\begin{planbox}{Temporal Scope Annotation Prompt}
\small
\textbf{System.} You are given the original question, the original ground truth answer, and the structured evidence JSON for one sample. Assign the single best \texttt{temporal\_scope} label. Use the question as the main reference.\\

\textbf{Label definitions.} \textsc{momentary}: a short instant is enough. \textsc{local window}: a short continuous span around the key event is enough. \textsc{long range}: evidence is spread across a large part of the clip. \textsc{multi segment}: several separated spans must be combined.\\

\textbf{Decision rule.} Choose the smallest temporal field that still supports the original answer.\\

\textbf{Output format.} Return valid JSON only in the form \texttt{\{"temporal\_scope": "LABEL"\}}.
\end{planbox}
}
\caption{\textbf{Temporal scope prompt,} assigning a temporal label from the structured evidence JSON.}
\label{fig:app_temporal_scope_prompt}
\end{figure}

\subsection{Reasoning Prompt and Consistency Filter}
\label{app:reason_prompt}

After the three task labels are predicted, Gemini-3.1-Pro generates the final tagged response from the clip, the original question, the original answer, the structured evidence JSON, and the predicted labels. The prompt asks for grounded explanation only and does not allow facts that are not already present in the JSON. The predicted labels are used to keep the reasoning aligned with the intended answer path. For example, \textsc{Visual} should focus on visual evidence, \textsc{temporal} should explain event order when needed, and \textsc{long range} or \textsc{multi segment} should connect cues across the required spans.

This stage also acts as a consistency filter. After generation, we normalize both the generated answer and the original answer by lowercasing, removing punctuation and articles, and collapsing spaces. If alias lists are available, we also accept a match to any alias. If the generated answer still does not match the original answer, we discard the sample. In this way, reasoning generation is used as a quality check rather than a new source of answer labels.

\begin{figure}[ht!]
\centering
\resizebox{1\textwidth}{!}{
\begin{planbox}{Reasoning Generation Prompt}
\small
\textbf{System.} You are given a video clip with audio, the original question, the original ground truth answer, the structured evidence JSON, and the predicted task labels. Generate the final tagged response.\\

\textbf{Input labels.} \texttt{evidence\_source}, \texttt{reasoning\_demand}, \texttt{temporal\_scope}.\\

\textbf{Output format.} Return exactly this format and nothing else: \texttt{<Reasoning>...</Reasoning><Answer>...</Answer>}.\\

\textbf{Reasoning rules.} Use only evidence that appears in the JSON. Keep the reasoning short, factual, and grounded. Mention only the cues that directly support the answer. Do not add new people, actions, emotions, or events that are not present in the JSON.\\

\textbf{Label-aware rules.} Follow the predicted labels when selecting cues. \textsc{Visual} should focus on visual evidence, \textsc{Audio} should focus on audio evidence, \textsc{Joint} should explain what comes from each modality, and \textsc{Conflict} should explain the mismatch. \textsc{temporal} should explain event order, \textsc{causal} should explain cause or effect, and \textsc{long range} or \textsc{multi segment} should connect evidence across the needed spans.\\

\textbf{Answer rule.} Keep \texttt{<Answer>} concise and semantically aligned with the original ground truth answer.
\end{planbox}
}
\caption{\textbf{Reasoning generation prompt,} generating the final tagged response conditioned on the structured evidence and the predicted task labels.}
\label{fig:app_reason_prompt}
\end{figure}

\subsection{Social Dimension Taxonomy and Prompt}
\label{app:social_dimension}

The field \texttt{social\_dimension} is used only in \benchmarknamenc{}. It is not part of the training target and is not used for expert routing. This field describes the main social phenomenon tested by the question. \textsc{Mental State Inference} is used for hidden emotion, belief, desire, uncertainty, or masked affect. \textsc{Pragmatic Meaning Inference} is used for indirect meaning, sarcasm, politeness, teasing, or relation signals in dialogue. \textsc{Action Goal Inference} is used for goal, plan, next step, or intention behind an action sequence. \textsc{Social Norm Inference} is used for role expectation, setting rule, social convention, or appropriateness in context. The same clip may receive different \texttt{social\_dimension} labels under different questions.

\begin{figure}[ht!]
\centering
\resizebox{1\textwidth}{!}{
\begin{planbox}{Social Dimension Annotation Prompt}
\small
\textbf{System.} You are given the original question, the original ground truth answer, and the structured evidence JSON for one benchmark sample. Assign the single best \texttt{social\_dimension} label. Use the question as the main reference.\\

\textbf{Label definitions.} \textsc{Mental State Inference}: hidden emotion, belief, desire, uncertainty, or other internal state. \textsc{Pragmatic Meaning Inference}: indirect meaning, sarcasm, teasing, politeness, power relation, or relation signal in dialogue. \textsc{Action Goal Inference}: goal, plan, next step, purpose, or intention behind an action sequence. \textsc{Social Norm Inference}: norm, role expectation, setting rule, appropriateness, or social convention.\\

\textbf{Decision rule.} Choose the single label that best captures the main kind of social reasoning tested by the question.\\

\textbf{Output format.} Return valid JSON only in the form \texttt{\{"social\_dimension": "LABEL"\}}.
\end{planbox}
}
\caption{\textbf{Social dimension prompt,} assigning a benchmark label from the structured evidence JSON.}
\label{fig:app_social_prompt}
\end{figure}

All samples in \benchmarknamenc{} then receive a final manual verification pass over the evidence JSON, the three task labels, the reasoning text, the final answer, and the \texttt{social\_dimension} field before release.

\section{\benchmarknamegradient{} Examples}
\label{app:benchexample}
\begin{figure}[ht!]
    \centering   
    \includegraphics[width=\linewidth]{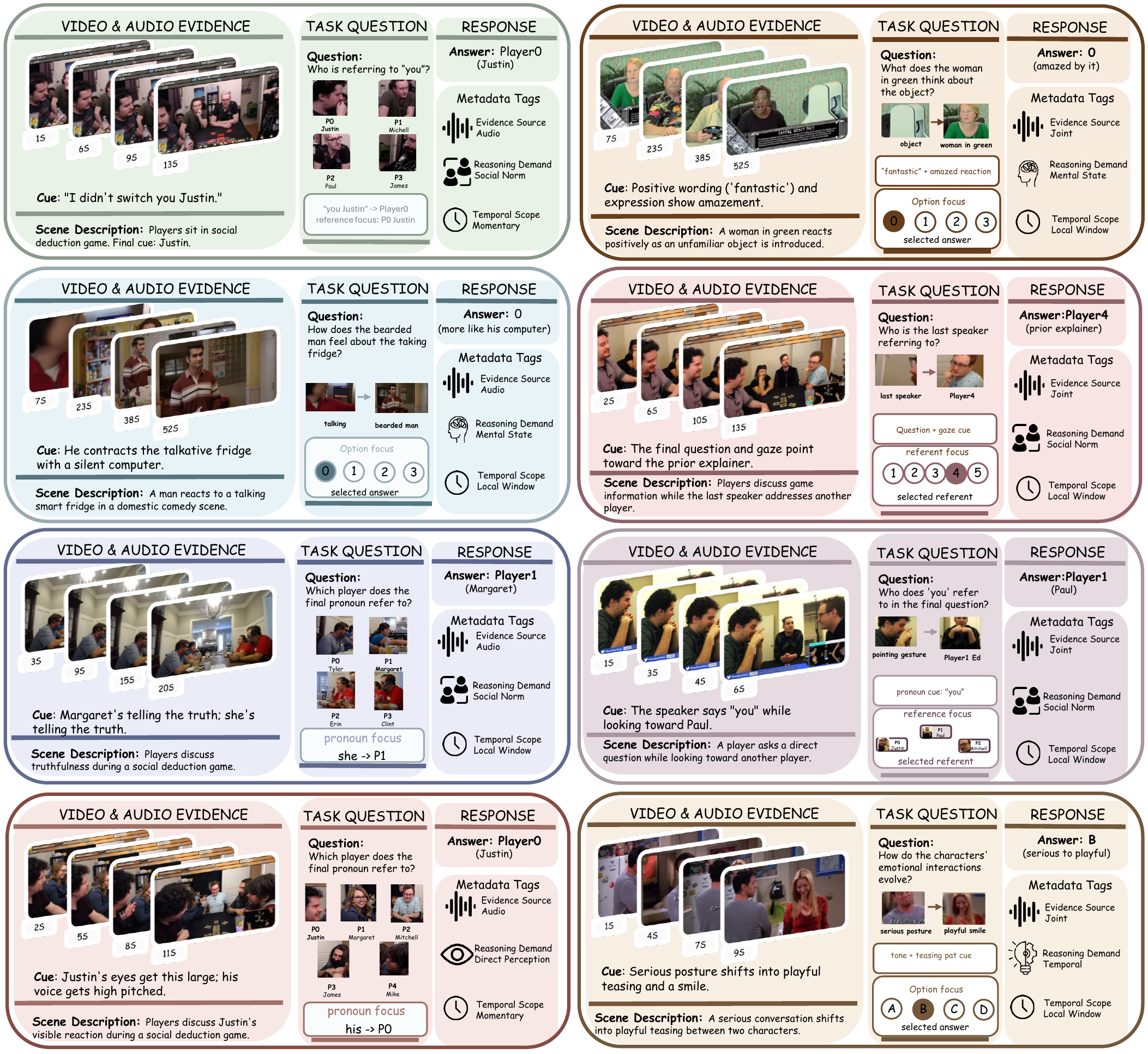}
    \caption{\textbf{Visualization of \benchmarknamenc{} examples,}
    showing representative social video QA instances with corresponding questions,
    answers, schema tags, grounded evidence, reasoning traces, and temporal evidence spans.}
    \label{fig:is_sample_qa_1}
\end{figure}

\begin{figure}[ht!]
    \centering    
    \includegraphics[width=\linewidth]{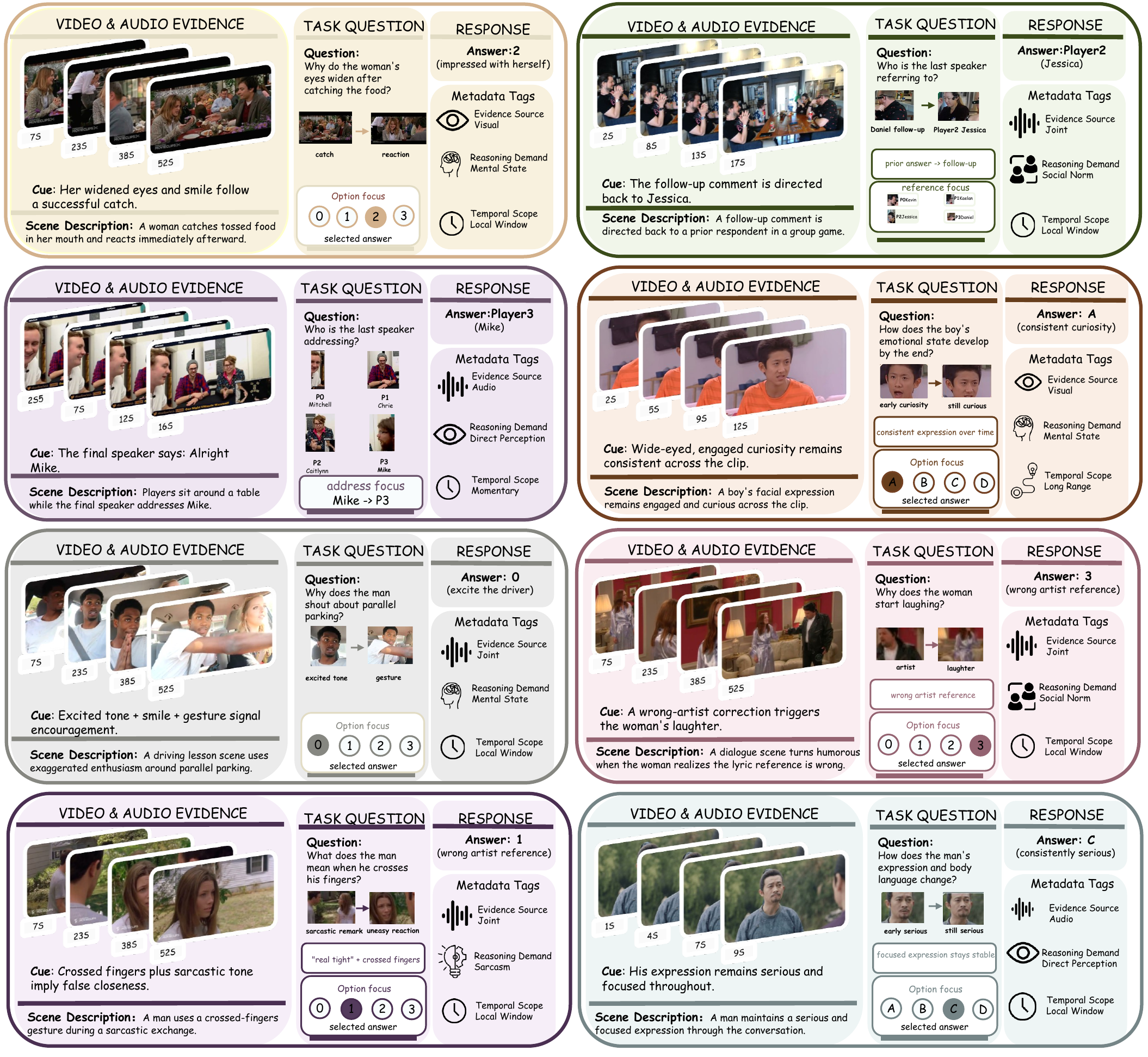}
    \caption{\textbf{Additional visualization of \benchmarknamenc{} examples.}
    These examples further illustrate the diversity of modality requirements, reasoning demands,
    and temporal scopes in \benchmarknamenc{}, including cases requiring audio-visual integration
    and socially grounded inference.\looseness-1}
    \label{fig:is_sample_qa_2}
\end{figure}

We provide qualitative examples from \benchmarknamenc{} to illustrate how each question is paired
with structured audio-visual evidence, schema labels, grounded reasoning, and temporal evidence
spans. As shown in Figures~\ref{fig:is_sample_qa_1} and~\ref{fig:is_sample_qa_2}, the benchmark
covers diverse social reasoning cases where the answer may rely on visual cues, audio cues, their
joint interpretation, or conflicts between what is said and what is visually expressed. These examples
also demonstrate how \benchmarknamenc{} makes the supporting evidence explicit, enabling diagnostic
evaluation beyond final-answer accuracy.

\section{Additional Experimental Details}
\subsection{Training Details}
\label{sec:training_detail}
For supervised fine-tuning (SFT), we train the base model, Qwen3 Omni 30B, on 8 NVIDIA H200 GPUs for one day, using a per-device batch size of 4 with a gradient accumulation step of 1. The learning rate is set to $1\times10^{-4}$, and we apply low-rank adaptation with rank 64. For reinforcement learning (RL), the loss coefficients are set to $\lambda_{\mathrm{ans}}=\lambda_{\mathrm{ctg}}=\lambda_{\mathrm{mcr}}=1$. We use AdamW optimization with standard $\beta_1=0.9$, $\beta_2=0.95$, and weight decay of 0.1, along with a cosine learning rate schedule and a short warmup phase. Mixed-precision training (bfloat16) is enabled for efficiency, and gradient clipping is applied to stabilize training. Additionally, we adopt a mixture-of-experts configuration with 128 total experts, from which 8 are selected per forward pass. To increase exploration and avoid over-reliance on the highest-scoring experts, we first identify the top 12 candidates according to the routing scores, and then select 8 experts from this subset, allowing more experts a chance to be utilized.

\subsection{Data Details}
Following the data construction and training paradigm detailed in Section~\ref{sec:sapr}, we adopt a staged pipeline with structured evidence extraction, schema annotation, and grounded reasoning generation to ensure high-quality supervision. Specifically, we construct a training corpus where each sample consists of multimodal inputs paired with structured evidence and schema-level annotations, and we retain only samples whose generated answers match the original ground-truth answers after normalization to reduce noise. For model training, we allocate 10k samples for training the lightweight MLP used in routing-related prediction components, 90k samples for the supervised fine-tuning (SFT) stage with schema-aligned objectives, and an additional 10k samples for the reinforcement learning stage, where both token generation and expert routing are optimized using outcome- and process-level rewards. For evaluation, we build a curated benchmark consisting of 800 samples spanning four representative tasks, each requiring different combinations of cross-modal reasoning, temporal grounding, and social inference. The evaluation split is selected by manual verification that the original answer is supported by observable audio-visual evidence. Gemini outputs are not used as correctness labels or as an inclusion criterion for the final benchmark split. Gemini was used only as an annotation assistant for training annotations; the final evaluation labels are human-verified labels. We further apply strict filtering, structured annotation, and manual verification on the benchmark split to guarantee consistency between evidence, reasoning, and final answers.

\section{Qualitative Response Visualization and Analysis}
\label{app:response}
We provide qualitative response comparisons between \modelnamenc{} and Gemini~3.1~Pro on manually verified examples from \benchmarknamenc{}. These examples are intended as diagnostic visualizations rather than an additional quantitative metric. Each case shows the input frames, question, candidate options, Gemini's response, and the correct response produced by \modelnamenc{}. The examples cover three common social reasoning settings: discourse-grounded reference resolution, multi-party turn-taking, and long-range affect continuity.\looseness-1

\begin{figure}[ht!]
    \centering
    \includegraphics[width=0.98\linewidth]{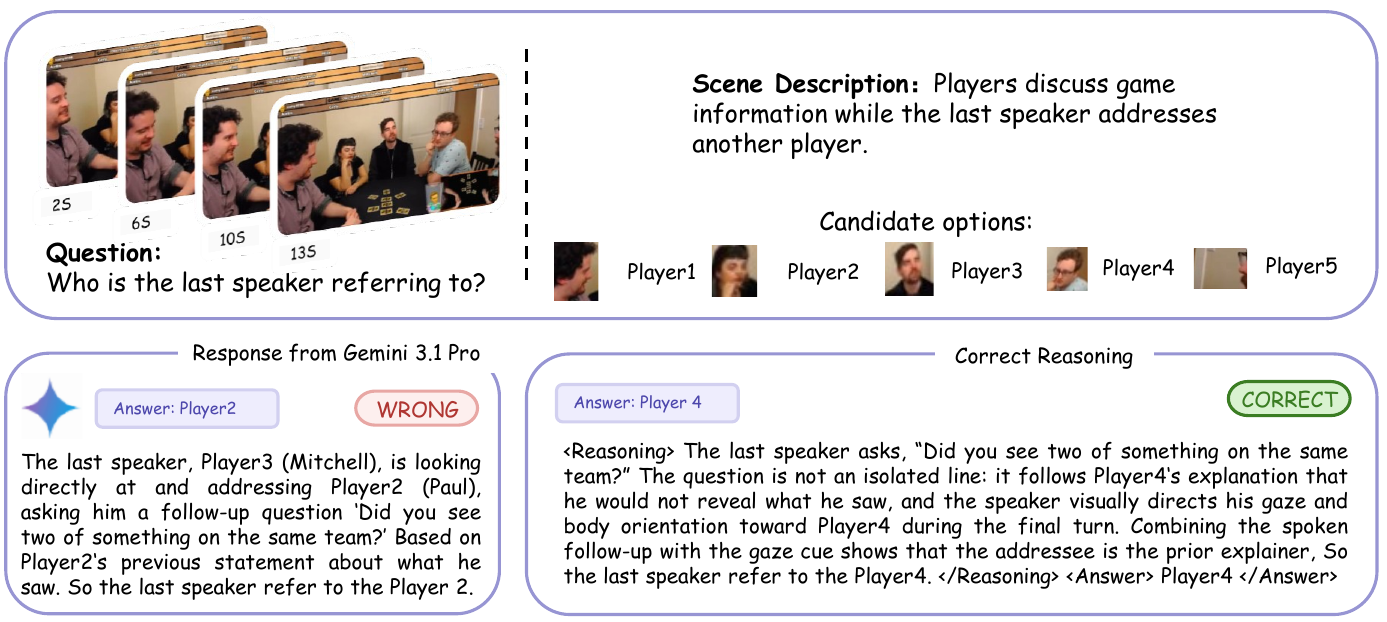}
    \caption{
    \textbf{Qualitative comparison on discourse-grounded reference resolution.}
    Gemini~3.1~Pro selects Player2 by treating the final utterance as a local follow-up to Player2's previous statement. \modelnamenc{} instead links the final question to Player4's prior explanation and uses the speaker's gaze and body orientation toward Player4, leading to the correct referent.
    }
    \label{fig:appendix_e_case1_reference}
\end{figure}

\begin{figure}[ht!]
    \centering
    \includegraphics[width=0.98\linewidth]{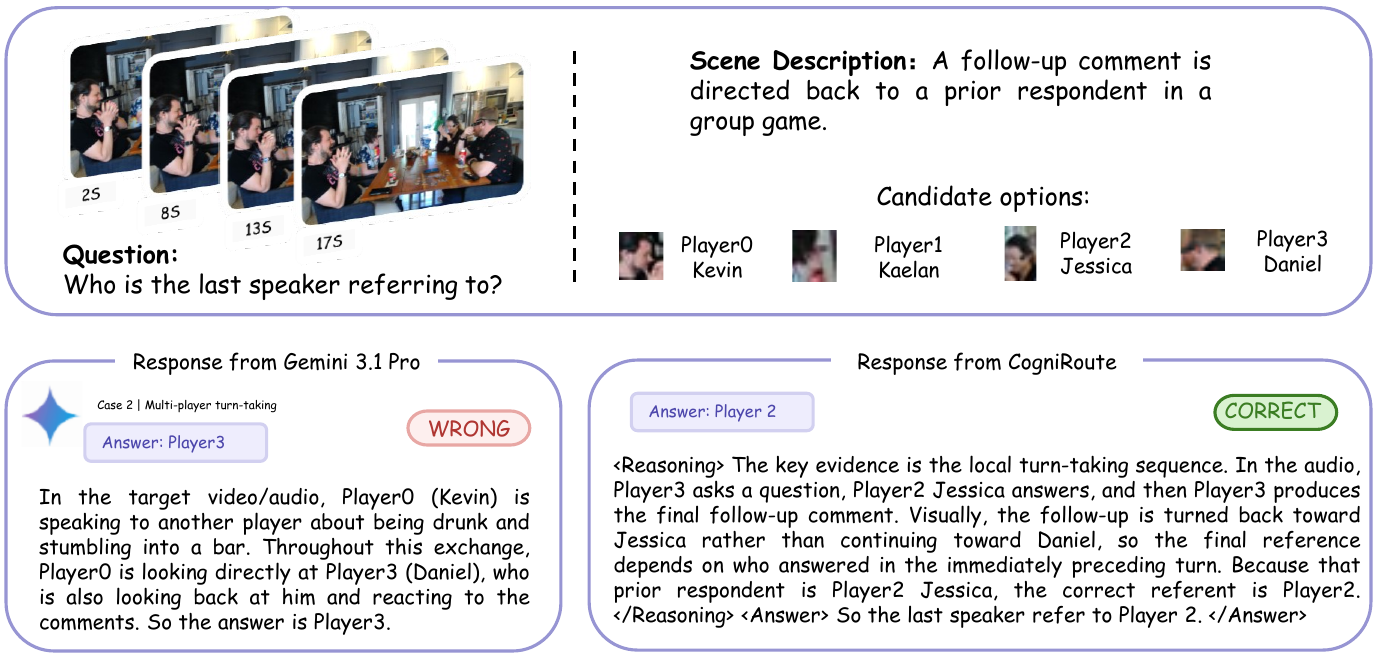}
    \caption{
    \textbf{Qualitative comparison on multi-party turn-taking.}
    Gemini~3.1~Pro over-relies on the visually salient gaze direction toward Player3 and predicts Player3 as the referent. CogniRoute follows the local dialogue sequence: Player3 asks a question, Player2 Jessica answers, and the subsequent follow-up is directed back to the prior respondent. This turn-taking evidence identifies Player2 as the correct referent.
    }
    \label{fig:appendix_e_case2_turntaking}
\end{figure}

\begin{figure}[ht!]
    \centering
    \includegraphics[width=0.98\linewidth]{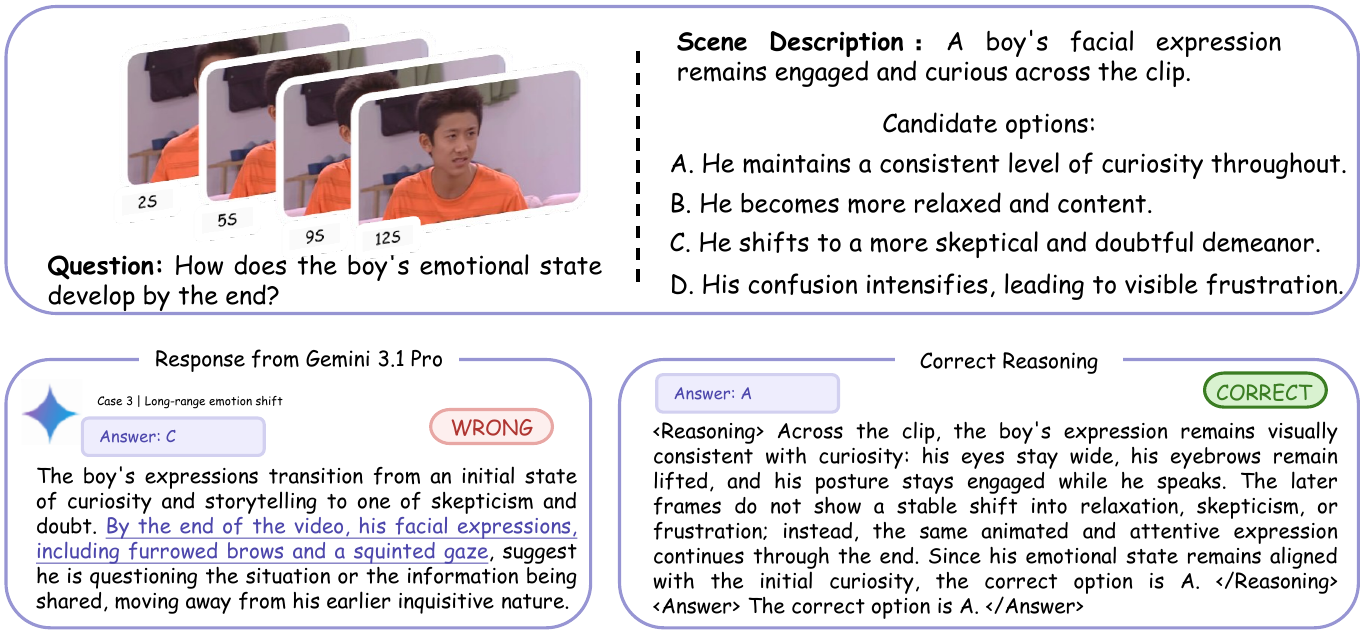}
    \caption{
    \textbf{Qualitative comparison on long-range affect grounding.}
    Gemini~3.1~Pro predicts an emotional shift toward skepticism or doubt from isolated late-frame cues. CogniRoute aggregates the boy's expression across the clip and observes that his wide-eyed, engaged expression remains consistent, supporting the correct answer that his curiosity is maintained throughout the video.
    }
    \label{fig:appendix_e_case3_emotion}
\end{figure}

Figures \ref{fig:appendix_e_case1_reference}-\ref{fig:appendix_e_case3_emotion} show Gemini~3.1~Pro often identifies a locally plausible cue, but maps it to the wrong evidence structure. In the first two cases, the answer depends on discourse-level reference tracking rather than a single utterance or a single gaze direction. In the third case, the answer depends on temporal continuity rather than a local facial expression. CogniRoute is better aligned with these requirements because its routing behavior is trained to reflect the schema axes of evidence source, reasoning demand, and temporal scope. The qualitative results therefore complement the quantitative gains in Table \ref{tab:benchmark_results} by illustrating how CogniRoute improves evidence selection, cross-modal grounding, and temporal reasoning in social video question answering.

\section{Additional Experiments and Ablations}\label{sec:ablationsmore}
\subsection{Training Generalization}
\begin{table}[t]
\centering
\caption{\textbf{Social understanding results} with a baseline and its retrained variant under identical settings. Numbers represent accuracy ($\uparrow$). Bold indicates the best result in each column.}
\resizebox{\textwidth}{!}{%
\begin{tabular}{lccccc}
\toprule
& Mental State & Pragmatic Meaning & Action Goal & Social Norm & Avg.\\
\midrule
OmniVinci~\cite{ye2025omnivinci}    & 26.07 & 21.07 & 20.81 & 18.75 & 21.88 \\
OmniVinci~\cite{ye2025omnivinci}(SFT)       & 46.92 & 43.21 & 47.65 & 39.38 & 44.25 \\
\midrule
\rowcolor{color4} CogniRoute (Ours) & \textbf{60.18}    & \textbf{55.71}    & \textbf{66.44}    & \textbf{58.13}    & \textbf{59.38}    \\
\bottomrule
\end{tabular}
}
\label{tab:sft_baseline}
\end{table}

To evaluate generalization under identical training settings, we compare CogniRoute with a baseline model OmniVinci and its retrained variant. As shown in Table \ref{tab:sft_baseline}, CogniRoute consistently achieves stronger performance across all four social dimension categories, and OmniVinci also gains a performance increase under the same training setting, indicating better transfer and generalization ability under the same supervision regime.\looseness-1

\subsection{SAPR Design}
\label{sec:ablation_sapr}

We evaluate the design of SAPR during supervised finetuning. All variants use the same backbone, data, training steps, and load balancing loss. The actual MoE computation always selects (8) experts from (128) experts. SAPR only changes the auxiliary routing signature used for the alignment loss. Full SAPR uses the correct schema tag set (C), the full (128) dimensional gate probability vector, modality-wise pooling, and a separate projection $P_l$ for each MoE layer. For the shuffled tag control, each sample is paired with the real schema tags of another sample, which keeps the global tag statistics but breaks the relation between the sample and its schema. For the random tag control, tags are sampled from the tag vocabulary while matching the original tag count. The shared projection variant uses one projection matrix for all MoE layers instead of layer-specific $P_l$. Since experts from different layers do not have the same meaning, this shared mapping loses layer information.

\begin{table}[!t]
\centering
\caption{\textbf{SAPR design ablation after supervised finetuning.} Numbers represent accuracy. MoE computation always selects (8) experts from (128) experts. Full SAPR uses the correct schema tags, the full (128) dimensional gate distribution, modality-wise pooling, and layer-specific projections.}
\label{tab:sapr_main_ablation}
\resizebox{\linewidth}{!}{%
\begin{tabular}{lccccc}
\toprule
Model Variant & Mental State & Pragmatic Meaning & Action Goal & Social Norm & Avg. \\ 
\midrule
Standard SFT without SAPR & 40.28 & 38.93 & 44.97 & 37.50 & 40.13 \\ 
SAPR with shuffled tags & 41.71 & 39.64 & 46.31 & 39.38 & 41.38 \\ 
SAPR with random tags & 40.76 & 38.93 & 45.64 & 38.75 & 40.63 \\ 
Shared projection across layers & 45.02 & 43.21 & 48.99 & 41.88 & 44.50 \\
\midrule
\rowcolor{color4} Full SAPR & \textbf{51.66} & \textbf{49.29} & \textbf{53.69} & \textbf{46.25} & \textbf{50.13} \\ 
\bottomrule
\end{tabular}%
}
\end{table}

As shown in Table \ref{tab:sapr_main_ablation}, Full SAPR gives the best result across all four categories. Shuffled tags and random tags are close to standard SFT, which shows that the gain does not come only from adding an auxiliary loss or extra parameters. The shared projection variant is also clearly lower than Full SAPR. This supports the need to keep layer information before mapping routing statistics into the shared schema space.

\subsection{Checking Learnable Tag Embedding Collapse}
\label{sec:ablation_tag_collapse}

SAPR uses a learnable tag embedding matrix \(W_{\mathrm{tag}}\). A possible issue is that different tag embeddings may collapse to similar directions. If this happens, the cosine alignment loss can be reduced without forcing the router to separate different cognitive schemas. We check this issue with two diagnostics. First, we measure the geometry of \(W_{\mathrm{tag}}\) by the mean pairwise cosine, the maximum pairwise cosine, and the normalized effective rank. Second, we test whether the routing signature \(g\) predicts the true schema tags by training a linear probe on frozen routing signatures. Standard SFT does not use \(W_{\mathrm{tag}}\), so it is used only as a routing baseline.\looseness-1

Let \(\sigma_r\) be the singular values of the row normalized tag embedding matrix, and let \(p_r=\sigma_r/\sum_s\sigma_s\). The normalized effective rank is defined as
\[
\mathrm{ERank}
=
\frac{
\exp\left(-\sum_r p_r\log p_r\right)
}{
\min(|\mathcal{V}_{\mathrm{tag}}|,D)
}.
\]
A value close to \(1\) means that the tag embedding matrix uses many independent directions, while a low value means that the embeddings are close to a low rank solution.

\begin{table}[!t]
\centering
\caption{\textbf{Diagnostics for learnable tag embedding collapse.} Route mAP is obtained by training a linear probe on frozen routing signatures to predict the true schema tags. Mean Cos. and Max Cos. are computed over all pairs of row-normalized tag embeddings. Norm. ERank is the normalized effective rank of \(W_{\mathrm{tag}}\). Standard SFT does not contain \(W_{\mathrm{tag}}\), so embedding geometry is not available.}
\label{tab:tag_embedding_collapse}
\resizebox{\linewidth}{!}{%
\begin{tabular}{lccccc}
\toprule
Model Variant & Route mAP & Mean Cos. & Max Cos. & Norm. ERank & Avg. Acc. \\
\midrule
Standard SFT without SAPR & 32.41 & N.A. & N.A. & N.A. & 40.13 \\
SAPR with shuffled tags & 34.18 & 0.09 & 0.34 & 0.87 & 41.38 \\
SAPR with random tags & 33.27 & 0.11 & 0.37 & 0.84 & 40.63 \\
\midrule
\rowcolor{color4} Full SAPR & \textbf{62.85} & \textbf{0.06} & \textbf{0.28} & \textbf{0.92} & \textbf{50.13} \\
\bottomrule
\end{tabular}%
}
\end{table}

As shown in Table \ref{tab:tag_embedding_collapse}, Full SAPR has low mean and maximum pairwise cosine values and a high normalized effective rank. This shows that the learnable tag embeddings do not collapse to the same direction. Full SAPR also has a much higher route mAP than the shuffled and random tag controls, which means that good embedding geometry alone is not enough. The correct relation between the sample and its schema tags is needed to make the routing signature schema aware.

\subsection{Token Branch Versus Gate Branch}
\label{sec:ablation_token_gate}

We study whether the gate branch in RMRL is needed. All variants start from the same SFT checkpoint trained with Full SAPR and use the same reward. Token branch only updates the language policy using the rollout advantage. The trainable router without gate objective allows router parameters to be updated during RL, but does not apply the route probability ratio. Gate branch only updates router related parameters while keeping the language path fixed. Full RMRL jointly optimizes token generation and route allocation.

\begin{table}[!t]
\centering
\caption{\textbf{Ablation of token and gate branches in RMRL.} All variants start from the same SFT checkpoint trained with Full SAPR and use the same reward. Numbers represent accuracy. Full RMRL performs best, showing that explicit gate optimization gives gains beyond token-level RL.}
\label{tab:token_gate_ablation}
\resizebox{\linewidth}{!}{%
\begin{tabular}{lccccc}
\toprule
Model Variant & Mental State & Pragmatic Meaning & Action Goal & Social Norm & Avg. \\
\midrule
SFT with Full SAPR & 51.66 & 49.29 & 53.69 & 46.25 & 50.13 \\
Token branch only & 57.36 & 52.86 & 62.42 & 54.38 & 56.13 \\
Token branch, trainable router, no gate objective & 57.82 & 53.21 & 63.09 & 55.63 &56.75 \\
Gate branch only &53.55 & 50.35 & 55.70 & 48.75 &51.88 \\
\midrule
\rowcolor{color4} Full RMRL & \textbf{60.18} & \textbf{55.71} & \textbf{66.44} & \textbf{58.13} & \textbf{59.38} \\
\bottomrule
\end{tabular}%
}
\end{table}

As shown in Table \ref{tab:token_gate_ablation}, token branch only already improves over the SFT checkpoint, which shows that outcome level reward helps token generation. However, Full RMRL further improves all four categories. The trainable router without gate objective is close to token branch only and remains clearly below Full RMRL. This shows that simply making the router trainable is not enough. The route probability ratio is needed to assign outcome level credit to expert selection. Gate branch only gives a smaller gain, which is expected because routing changes alone cannot fully improve the generated reasoning. These results support the joint optimization of token generation and expert allocation.\looseness-1

\subsection{Routing Behavior Under Gate Optimization}
\label{sec:ablation_gate_metrics}

We further analyze how the gate branch changes routing behavior. Accuracy alone cannot show whether the gain comes from better expert allocation. We therefore report routing metrics that test three properties. Route mAP measures whether routing signatures predict the schema tags. Load CV and Active Exp. measure whether the model keeps a balanced use of experts. The two route shift metrics measure whether routing changes more when the required modality is removed than when an irrelevant modality is removed.

For this analysis, routing statistics are computed on generated reasoning tokens, since the gate branch receives RL feedback through generated token routes. Route mAP is obtained by training a linear probe on frozen routing signatures to predict the true schema tags. Load CV is the coefficient of variation of selected expert counts over the \(128\) experts. Active Exp. is the number of experts that receive at least \(0.1\%\) of selected routes. To compute modality route shift, we first generate the output with the full input and then run teacher forcing on the same output under masked inputs. Let \(g_{\mathrm{gen}}\) denote the routing signature over generated reasoning tokens and let \(d(\cdot,\cdot)\) be cosine distance. We define
\[
\Delta_{\mathrm{rel}}
=
d\left(g_{\mathrm{gen}}(x), g_{\mathrm{gen}}(x_{\setminus m_{\mathrm{rel}}})\right),
\qquad
\Delta_{\mathrm{irrel}}
=
d\left(g_{\mathrm{gen}}(x), g_{\mathrm{gen}}(x_{\setminus m_{\mathrm{irrel}}})\right).
\]
For visual only samples, video is the relevant modality and audio is the irrelevant modality. For audio only samples, audio is the relevant modality and video is the irrelevant modality. For samples that require both audio and video, removing either modality is treated as relevant and these samples are not used for \(\Delta_{\mathrm{irrel}}\).

\begin{table}[!t]
\centering
\caption{\textbf{Routing behavior under gate optimization.} Route mAP is measured by a linear probe on frozen routing signatures. Load CV and Active Exp. measure expert balance and expert coverage. \(\Delta_{\mathrm{rel}}\) and \(\Delta_{\mathrm{irrel}}\) measure route changes after masking the required and irrelevant modalities. Full RMRL improves schema-aware and modality-aware routing while keeping expert use balanced.}
\resizebox{\linewidth}{!}{%
\begin{tabular}{lcccccc}
\toprule
Model Variant & Avg. Acc. \(\uparrow\) & Route mAP \(\uparrow\) & Load CV \(\downarrow\) & Active Exp. \(\uparrow\) & \(\Delta_{\mathrm{rel}}\uparrow\) & \(\Delta_{\mathrm{irrel}}\downarrow\) \\
\midrule
SFT with Full SAPR & 50.13 & 62.85 & 0.26 & 116 & 0.141 & 0.090 \\
Token branch only & 56.13 & 64.07 & 0.27 & 114 & 0.154 & 0.095 \\
Token branch, trainable router, no gate objective & 56.75 & 65.21 & 0.29 & 111 & 0.163 & 0.101 \\
Gate branch only & 51.88 & 67.48 & 0.24 & 120 & 0.181 & 0.087 \\
\midrule
\rowcolor{color4} Full RMRL & \textbf{59.38} & \textbf{73.92} & \textbf{0.22} & \textbf{124} & \textbf{0.224} & \textbf{0.082} \\
\bottomrule
\end{tabular}%
}
\label{tab:gate_routing_metrics}
\end{table}

As shown in Table \ref{tab:gate_routing_metrics}, Full RMRL gives the highest Route mAP, which shows that gate optimization makes generated token routing more predictive of the schema tags. It also gives the largest \(\Delta_{\mathrm{rel}}\) and the smallest \(\Delta_{\mathrm{irrel}}\), which means that routing changes more when the evidence source is removed and changes less when an unrelated modality is removed. The Load CV and Active Exp. results show that this gain is not caused by routing collapse. The trainable router without the gate objective is clearly weaker than Full RMRL, which again shows that making the router trainable is not enough. The route probability ratio is needed to assign outcome-level credit to expert selection.

\subsection{SFT and RL Rewards}
\begin{table}[t]
\centering
\caption{\textbf{Ablation study evaluating the impact of SFT and RL rewards on benchmark performance.} $\mathcal{R}_{ctg}$ denotes the Cognitive Temporal Grounding reward, $\mathcal{R}_{mcr}$ denotes the Modality-Consistent Reasoning reward, and $\mathcal{R}_{ans}$ denotes the Answer Correctness reward. Numbers represent accuracy ($\uparrow$). Bold indicates the best result in each column.}
\resizebox{0.85\textwidth}{!}{%
\begin{tabular}{lccccc}
\toprule
Model Variant & Mental State & Pragmatic Meaning & Action Goal & Social Norm & Avg. \\
\midrule
SFT & 51.66 & 49.29 & 53.69 & 46.25 & 50.13 \\
SFT + RL w/o $\mathcal{R}_{ctg}$ & 54.98 & 51.43 & 57.72 & 51.25 & 53.50 \\
SFT + RL w/o $\mathcal{R}_{mcr}$ & 56.40 & 53.21 & 60.40 & 53.13 & 55.38 \\
SFT + RL w/o $\mathcal{R}_{ans}$ & 57.82 & 54.64 & 62.42 & 55.63 & 57.13 \\
\midrule
\rowcolor{color4} SFT + Full RL (Ours) & \textbf{60.18} & \textbf{55.71} & \textbf{66.44} & \textbf{58.13} & \textbf{59.38} \\
\bottomrule
\end{tabular}%
}
\label{tab:rl_ablation}
\end{table}
To understand the contribution of each component in our RMRL reinforcement learning stage, we conduct an ablation study on the proposed reward functions. As shown in Table \ref{tab:rl_ablation}, applying only SFT establishes a solid baseline, but performance improves significantly when reinforcement learning is introduced. We systematically remove each reward to isolate its effects. The results demonstrate that dropping any single reward degrades performance across all social dimensions.

\subsection{CTG Evaluation Beyond Its Reward Value}
\label{sec:ablation_ctg_eval}

We evaluate whether CTG improves temporal grounding beyond increasing its own reward value. Our evidence annotation provides one main evidence time for each sample, rather than a start and end boundary. Therefore, we do not report temporal IoU. Instead, we compare the model's final reasoning state with the annotated evidence time.

For all models, we compute the temporal attention distribution in the same way as CTG in Section~\ref{sec:ctg}. We use the short suffix of the reasoning span, including the closing reasoning token, average attention over the selected deep layers and heads, and pool audio and video tokens by temporal ID. We then evaluate whether this attention is close to the annotated time. Mass@0.5s and Mass@1.0s measure the attention mass within 0.5 seconds and 1.0 seconds of the annotated time. Hit@0.5s and Hit@1.0s measure whether the most attended time bin falls inside the same windows. PeakErr is the average absolute distance between the most attended time and the annotated time.

We also include a hard point CTG reward variant. This variant replaces our smooth CTG reward with a binary reward. It receives 1 if the most attended time bin from the reasoning suffix falls within 0.5 seconds of the annotated time, and 0 otherwise. Invalid reasoning suffixes also receive 0. Since each reward component is bounded by 1, this variant uses the same reward scale as our CTG. The full CTG row treats the annotated time as a zero length interval and applies the Gaussian proximity kernel from Section~\ref{sec:ctg}, which gives partial credit to near misses.\looseness-1

\begin{table}[!t]
\centering
\caption{\textbf{CTG evaluation with single point evidence annotations.} Mass and Hit reported in percentages, PeakErr in seconds. Temporal attention distribution computed from the reasoning suffix used by CTG. Full RMRL gives the best point-based temporal grounding results.}
\label{tab:ctg_point_eval}
\resizebox{\linewidth}{!}{%
\begin{tabular}{lcccccc}
\toprule
Model Variant & Avg. Acc. & Mass@0.5s \(\uparrow\) & Mass@1.0s \(\uparrow\) & Hit@0.5s \(\uparrow\) & Hit@1.0s \(\uparrow\) & PeakErr \(\downarrow\) \\
\midrule
SFT with Full SAPR & 50.13 & 15.6 & 26.8 & 16.3 & 28.1 & 2.94 \\
Full RL without \(\mathcal{R}_{\mathrm{ctg}}\) & 53.50 & 17.4 & 30.2 & 18.5 & 31.6 & 2.67 \\
Hard point CTG reward & 58.37 & 27.8 & 44.0 & 29.1 & 45.3 & 1.98 \\
Full RL with shuffled CTG time labels & 53.25 & 16.9 & 29.1 & 17.8 & 30.4 & 2.73 \\
\midrule
\rowcolor{color4} Full RMRL & \textbf{59.38} & \textbf{31.4} & \textbf{49.2} & \textbf{33.1} & \textbf{51.0} & \textbf{1.56} \\
\bottomrule
\end{tabular}%
}
\end{table}

As shown in Table~\ref{tab:ctg_point_eval}, Full RMRL improves both answer accuracy and point based temporal grounding. The gains hold under both 0.5 second and 1.0 second tolerance windows, and PeakErr is also lower. The hard point CTG reward improves over the model without \(\mathcal{R}_{\mathrm{ctg}}\), which shows that direct temporal supervision is useful. However, it is weaker than the full CTG reward because the binary score gives no partial credit to near misses and is more sensitive to small annotation offsets. The shuffled time label control is close to the model without \(\mathcal{R}_{\mathrm{ctg}}\), showing that the gain comes from the correct evidence time rather than from adding an extra reward term.\looseness-1

\subsection{Causal Evidence Deletion}
\label{sec:ablation_evidence_deletion}

We apply this test only to the final Full RMRL model, because the goal is to check input level evidence use of our final system. For each sample, we compare the original input, an input with the annotated evidence window removed, and an input with an equal-length random window removed. Since our evidence label is a single time point ($t_{\text{gt}}$), GT Mask removes all audio and video tokens whose temporal ids fall inside $[t_{\text{gt}}-0.5\text{s}, t_{\text{gt}}+0.5\text{s}]$. Question tokens are kept unchanged. Random Mask removes a window with the same length from the same clip, but outside the GT deletion window. For Random Mask, we sample five random windows for each sample and report the mean accuracy. All settings use the same decoding strategy as the main evaluation.\looseness-1

\begin{table}[!t]
\centering
\caption{\textbf{Causal evidence deletion test on the final Full RMRL model.} GT Mask removes the audiovisual window centered at the annotated evidence time. Random Mask removes an equal-length random audiovisual window from the same clip. Numbers represent accuracy.}
\label{tab:evidence_deletion_final}
\resizebox{0.85\linewidth}{!}{%
\begin{tabular}{lccccc}
\toprule
Input Variant & Mental State & Pragmatic Meaning & Action Goal & Social Norm & Avg. \\
\midrule
GT Mask & 47.87 & 43.21 & 53.69 & 44.38 & 46.63 \\
Random Mask & 57.35 & 52.14 & 62.42 & 55.63 & 56.13 \\
\midrule
\rowcolor{color4} Full Input & \textbf{60.18} &\textbf{55.71} & \textbf{66.44} & \textbf{58.13} & \textbf{59.38} \\
\bottomrule
\end{tabular}%
}
\end{table}

As shown in Table~\ref{tab:evidence_deletion_final}, removing the annotated evidence window causes a clear accuracy decrease across all four categories. In contrast, removing a random window with the same length has a much smaller effect. The average accuracy is (46.63) under GT Mask and (56.13) under Random Mask. This result indicates that the final model is more sensitive to the annotated evidence time than to random audiovisual content, which supports that the model uses the temporal evidence region when making its prediction.

\subsection{Candidate Expert Pool Size for Route Exploration}
\label{sec:ablation_topm}

Our MoE model has 128 experts and activates $k\!=\!8$ experts for each routed token. During reinforcement learning, we sample the eight active experts from the top $M$ router candidates. When $M!=\!8$, routing is the deterministic top 8 case. Increasing $M$ allows more route search, but very large $M$ may add low probability experts and make routing less stable. We sweep $M$ from 8 to 16.

\begin{table}[!t]
\centering
\caption{\textbf{Ablation of the candidate expert pool size (M) during RL route sampling.} 
M=12 yields best performance. Larger pools reduce accuracy, as lower probability experts add noisy route actions.}
\label{tab:topm_ablation}
\resizebox{\linewidth}{!}{%
\begin{tabular}{lccccccccc}
\toprule
Candidate Pool ($M$) & 8 & 9 & 10 & 11 & 12 & 13 & 14 & 15 & 16 \\
\midrule
Avg. Acc. & 58.00 & 58.50 & 58.88 & 59.13 & \textbf{59.38} & \underline{59.25} & 59.13 & 58.63 & 57.75 \\
\bottomrule
\end{tabular}%
}
\end{table}

The best result is obtained at $M=12$, with $M=13$ giving a very close score. We use $M=12$ in the final model because it gives the best accuracy. The deterministic case $M=8$ is weaker, which shows that route search is useful during reinforcement learning. Larger values such as $M=15$ and $M=16$ reduce accuracy, suggesting that too much route search can introduce noisy expert choices.

\section{Additional Related Work}

\noindent \textbf{Reinforced Multimodal Reasoning.}
Recent advances in Reinforcement Learning (RL), particularly through process supervision and verifiable reward modeling in language models~\cite{lightman2023let,guo2025deepseek,wang2025context}, have begun to extend into Multimodal Large Language Models (MLLMs), enabling improved reasoning capabilities across multimodal tasks~\cite{cao2026groundr1incentivizinggroundedvisual, duan2025gotr1unleashingreasoningcapability, su2025openthinkimglearningthinkimages, wang2025vgrefinertoolrefinedreferringgrounded, yang2025r1onevisionadvancinggeneralizedmultimodal, zhou2025r1zerosahamomentvisual, shen2025vlmr1stablegeneralizabler1style, fan2025gritteachingmllmsthink, zheng2026deepeyesincentivizingthinkingimages,shen2026fine}. These approaches introduce reward signals that supervise not only final predictions but also intermediate reasoning steps, often by verifying consistency with visual evidence through process reward models~\cite{wang2025visualprmeffectiveprocessreward, ong2025trainingvisionlanguageprocessreward,zhang2025gm,202606.0173,liu2026palm}. Building on R1-style training paradigms and RL algorithms such as GRPO~\cite{shao2024deepseekmathpushinglimitsmathematical}, recent works further encourage more structured reasoning trajectories to achieve grounded visual reasoning~\cite{yang2025r1onevisionadvancinggeneralizedmultimodal, shen2025vlmr1stablegeneralizabler1style, zhou2025r1zerosahamomentvisual, cao2026groundr1incentivizinggroundedvisual, chu2025qwenlookagainguiding,cao2024visual,li2026toward}. In parallel, methods like GRIT~\cite{fan2025gritteachingmllmsthink}, DeepEyes~\cite{zheng2026deepeyesincentivizingthinkingimages}, and related frameworks~\cite{wang2026traceableevidenceenhancedvisual, jiang2026embed} incorporate explicit visual grounding, \eg bounding boxes or region-level supervision, into the reasoning process. This alignment between textual inference and localized visual evidence improves interpretability and helps reduce hallucination~\cite{jiang2026embed}.\looseness-1

\textbf{Social omni-modal reasoning.}
Recent work has begun to move omni-modal modeling toward human-centered and socially interactive understanding. HumanOmni~\citep{zhao2025humanomni} develops a vision-speech-language model for human-centric video understanding, with emphasis on emotion recognition, facial-expression description, and action understanding. R1-Omni~\citep{zhao2025r1} applies reinforcement learning with verifiable rewards to omni-modal emotion recognition, improving reasoning, accuracy, and generalization. HumanOmniV2~\citep{yang2025humanomniv2} further studies context-aware omni-modal reasoning over human intentions and emotions, addressing global-context misunderstanding and shortcut behavior through context and logical rewards. Complementary benchmarks examine social interaction from more targeted perspectives: SocialOmni~\citep{xie2026socialomni} evaluates conversational interactivity through speaker identification, interruption timing, and interruption generation, while Omni-MMSI~\citep{li2026omni} focuses on identity-attributed social interaction understanding from raw audio, vision, and speech cues. These works establish social omni-modal understanding as an important direction, but they primarily focus on improving or evaluating high-level social perception, emotion/intent reasoning, turn-taking, or identity attribution. In contrast, our work treats social video question answering as an evidence-allocation problem. CogniRoute explicitly supervises the model's internal routing behavior according to the evidence structure of each question, including modality relation, reasoning demand, and temporal scope, and further optimizes expert allocation with rewards for answer correctness, modality-consistent reasoning, and temporal grounding.\looseness-1

\section{Broader Impacts}
\label{app:broader}
This work aims to improve omni-modal social video understanding by encouraging models to use appropriate visual, audio, and temporal evidence rather than relying only on final-answer supervision. The proposed dataset annotations and routing-aware training framework may benefit applications such as assistive video understanding, educational analysis, human-computer interaction, and socially aware dialogue systems. At the same time, social video understanding involves sensitive human behaviors, including speech, facial expressions, emotions, and interpersonal interactions. If deployed without safeguards, such systems could raise privacy concerns, amplify social or cultural biases, or be misused for surveillance and automated social profiling. We therefore emphasize that the proposed benchmark is intended for research evaluation, not for high-stakes decision making. Practical deployment should require consent-aware data collection, privacy protection, bias auditing across demographic and cultural groups, and clear restrictions against using model outputs as definitive judgments of a person’s intent, emotion, or social character.

\section{Limitations}
\label{app:limit}
Our work focuses on omni-modal social video question answering with pre-segmented clips and text-form answers. As a result, we do not evaluate online streaming interaction, real-time dialogue management, or speech generation quality. In addition, our Cognitive Schema uses a compact label space designed for routing supervision and analysis; extending it to more fine-grained social categories may be useful for domain-specific applications. Finally, the current implementation is based on an MoE omni backbone, so adapting the method to other architectures may require minor engineering changes. These limitations do not affect the core conclusion that sample-level schema guidance can improve expert routing and audio-visual reasoning.

\end{document}